# An Investigation into Mathematical Programming for Finite Horizon Decentralized POMDPs

**Raghav Aras** RAGHAV.ARAS@GMAIL.COM
*IMS, Suplec Metz*
*2 rue Edouard Bélin, Metz Technopole*
*57070 Metz - France*

**Alain Dutech** ALAIN.DUTECH@LORIA.FR
*MAIA - LORIA/INRIA*
*Campus Scientifique - BP 239*
*54506 Vandoeuvre les Nancy - France*

## Abstract

Decentralized planning in uncertain environments is a complex task generally dealt with by using a decision-theoretic approach, mainly through the framework of Decentralized Partially Observable Markov Decision Processes (DEC-POMDPs). Although DEC-POMDPS are a general and powerful modeling tool, solving them is a task with an overwhelming complexity that can be doubly exponential. In this paper, we study an alternate formulation of DEC-POMDPs relying on a *sequence-form* representation of policies. From this formulation, we show how to derive Mixed Integer Linear Programming (MILP) problems that, once solved, give *exact* optimal solutions to the DEC-POMDPs. We show that these MILPs can be derived either by using some combinatorial characteristics of the optimal solutions of the DEC-POMDPs or by using concepts borrowed from game theory. Through an experimental validation on classical test problems from the DEC-POMDP literature, we compare our approach to existing algorithms. Results show that mathematical programming outperforms dynamic programming but is less efficient than forward search, except for some particular problems.

The main contributions of this work are the use of *mathematical programming* for DEC-POMDPs and a better understanding of DEC-POMDPs and of their solutions. Besides, we argue that our alternate representation of DEC-POMDPs could be helpful for designing novel algorithms looking for approximate solutions to DEC-POMDPs.

## 1. Introduction

The framework of Decentralized Partially Observable Markov Decision Processes (DEC-POMDPs) can be used to model the problem of designing a system made of autonomous agents that need to coordinate in order to achieve a joint goal. Solving DEC-POMDPs is an untractable task as they belong to the class of NEXP-complete problems (see Section 1.1). In this paper, DEC-POMDPs are reformulated into *sequence-form DEC-POMDPs* so as to derive Mixed Integer Linear Programs that can be solved using very efficient solvers in order to design *exact* optimal solutions to finite-horizon DEC-POMDPs. Our main motivation is to investigate the benefits and limits of this novel approach and to get a better understanding of DEC-POMDPs (see Section 1.2). On a practical level, we provide new algorithms and heuristics for solving DEC-POMDPs and evaluate them on classical problems (see Section 1.3).

### 1.1 Context

One of the main goals of Artificial Intelligence is to build artificial agents that exhibit intelligent behavior. An agent is an entity situated in an environment which it can perceive through sensors and act upon using actuators. The concept of planning, *i.e.*, to select a sequence of actions in order to reach a goal, has been central to the field of Artificial Intelligence for years. While the notion of "intelligent behavior" is difficult to assess and to measure, we prefer to refer to the concept of "rational behavior" as formulated by Russell and Norvig (1995). As a consequence, the work presented here uses a decision-theoretic approach in order to build agents that take optimal actions in an uncertain and partially unknown environment.

We are more particularly interested in cooperative multi-agent systems where multiple independent agents with limited perception of their environment must interact and coordinate in order to achieve a joint task. No central process with a full knowledge of the state of the system is there to control the agents. On the contrary, each agent is an autonomous entity that must execute its actions by itself. This setting is both a blessing, as each agent should ideally deal with a small part of the problem, and a curse, as coordination and cooperation are harder to develop and to enforce.

The decision-theoretic approach to rational behavior relies mostly on the framework of Markov Decision Processes (MDP) (Puterman, 1994). A system is seen as a sequence of discrete states with stochastic dynamics, some particular states giving a positive or negative reward. The process is divided into discrete decision periods; the number of such periods is called the *horizon* of the MDP. At each of these periods, an action is chosen which will influence the transition of the process to its next state. By using the right actions to influence the transition probabilities between states, the objective of the controller of the system is to maximize its long term return, which is often an additive function of the reward earned for the given horizon. If the controller knows the dynamics of the system, which is made of a transition function and of a reward function, algorithms derived from the field of Dynamic Programming (see Bellman, 1957) allow the controller to compute an optimal deterministic *policy*, *i.e.*, a decision function which associates an "optimal" action to every state so that the expected long term return is optimal. This process is called *planning* in the MDP community.

In fact, using the MDP framework, it is quite straightforward to model a problem with one agent which has a full and complete knowledge of the state of the system. But agents, and especially in a multi-agent setting, are generally not able to determine the complete and exact state of the system because of noisy, faulty or limited sensors or because of the nature of the problem itself. As a consequence, different states of the system are observed as similar by the agent which is a problem when different optimal actions should be taken in these states; one speaks then of *perceptual aliasing*. An extension of MDPs called Partially Observable Markov Decisions Processes (POMDPs) deals explicitly with this phenomenon and allows a single agent to compute plans in such a setting provided it knows the conditional probabilities of observations given the state of the environment (Cassandra, Kaelbling, & Littman, 1994).

As pointed out by Boutilier (1996), multi-agent problems could be solved as MDPs if considered from a centralized point of view for planning *and* control. Here, although planning is a centralized process, we are interested in decentralized settings where every agent executes its own policy. Even if the agents could instantly communicate their observation, we consider problems where the joint observation resulting from such communications would still not be enough to identify the state of the system. The framework of Decentralized Partially Observable Markov Decision Processes (DEC-POMDP) proposed by Bernstein, Givan, Immerman, and Zilberstein (2002) takes into account decentralization of control and partial observability. In a DEC-POMDP, we are looking for optimal *joint policies* which are composed of one policy for each agent, these individual policies being computed in a centralized way but then independently executed by the agents.

The main limitation of DEC-POMDPs is that they are provably untractable as they belong to the class of NEXP-complete problems (Bernstein et al., 2002). Concretely, this complexity result implies that, in the worst case, finding an optimal joint policy of a finite horizon DEC-POMDP requires time that is *exponential* in the horizon if one always make good choices. Because of this complexity, there are very few algorithms for finding exact optimal solutions for DEC-POMDPs (they all have a doubly exponential complexity) and only a few more that look for approximate solutions. As discussed and detailed in the work of Oliehoek, Spaan, and Vlassis (2008), these algorithms follow either a dynamic programming approach or a forward search approach by adapting concepts and algorithms that were designed for POMDPs.

Yet, the concept of decentralized planning has been the focus of quite a large body of previous work in other fields of research. For example, the *Team Decision Problem* (Radner, 1959), later formulated as a Markov system in the field of control theory by Anderson and Moore (1980), led to the *Markov Team Decision Problem* (Pynadath & Tambe, 2002). In the field of mathematics, the abundant literature on Game Theory brings a new way for looking at multi-agent planning. In particular, a DEC-POMDP with finite horizon can be thought as a *game in extensive form with imperfect information and identical interests* (Osborne & Rubinstein, 1994).

Taking inspiration from the field of game theory and mathematical programming to design exact algorithms for solving DEC-POMDPs is precisely the subject of our contribution to the field of decentralized multi-agent planning.

### 1.2 Motivations

The main objective of our work is to investigate the use of mathematical programming, more especially mixed-integer linear programs (MILP) (Diwekar, 2008), for solving DEC-POMDPs. Our motivation relies on the fact that the field of linear programming is quite mature and of great interest to the industry. As a consequence, there exist many efficient solvers for mixed-integer linear programs and we want to see how these efficient solvers perform in the framework of DEC-POMDPs.

Therefore, we have to reformulate a DEC-POMDP to solve it as a mixed-integer linear program. As shown in this article, two paths lead to such mathematical programs, one grounded on the work from Koller, Megiddo, and von Stengel (1994), Koller and Megiddo (1996) and von Stengel (2002), and another one grounded on combinatorial considerations. Both methods rely on a special reformulation of DEC-POMDPs in what we have called *sequence-form DEC-POMDPs* where a policy is defined by the histories (*i.e.*, sequences of observations and actions) it can generate when applied to the DEC-POMDP.

The basic idea of our work is to select, among all the histories of the DEC-POMDP, the histories that will be part of the optimal policy. To that end, an optimal solution to the MILP presented in this article will assign a positive weight to each history of the DEC-POMDP and every history with a non-negative weight will be part of the optimal policy to the DEC-POMDP. As the number of possible histories is exponential in the horizon of the problem, the complexity of a naive search for the optimal set of histories is doubly exponential. Therefore, our idea appears untractable and useless.

Nevertheless, we will show that combining the efficiency of MILP solvers with some quite simple heuristics leads to exact algorithms that compare quite well to some existing exact algorithms. In fact, sequence-form DEC-POMDPs only need a memory space exponential in the size of the problem. Even if solving MILPs can also be exponential in the size of the MILP and thus leads to doubly exponential complexity for sequence-form based algorithms, we argue that sequence-form MILPs compare quite well to dynamic programming thanks to optimized industrial MILP solvers like “Cplex”.

Still, our investigations and experiments with Mathematical Programming for DEC-POMDPs do not solely aim at finding exact solutions to DEC-POMDPs. Our main motivation is to have a better understanding of DEC-POMDPs and of the limits and benefits of the mathematical programming approach. We hope that this knowledge will help deciding to what extent mathematical programming and sequence-form DEC-POMDPs can be used to design novel algorithms that look for *approximate* solutions to DEC-POMDPs.

### 1.3 Contributions

In this paper we develop new algorithms in order to find exact optimal joint policies for DEC-POMDPs. Our main inspiration comes from the work of Koller, von Stegel and Megiddo that shows how to solve games in extensive form with imperfect information and identical interests, that is how to find a *Nash equilibrium* for this kind of game (Koller et al., 1994; Koller & Megiddo, 1996; von Stengel, 2002). Their algorithms caused a breakthrough as the memory space requirement of their approach is linear in the size of the game whereas more canonical algorithms required space that is exponential in the size of the game. This breakthrough is mostly due to the use of a new formulation of a policy in what they call a *sequence-form*.

Our main contribution, as detailed in Section 3.3, is then to adapt the **sequence-form** introduced by Koller, von Stegel and Megiddo to the framework of DEC-POMDPs (Koller et al., 1994; Koller & Megiddo, 1996; von Stengel, 2002). As a result, it is possible to formulate the resolution of a DEC-POMDP as a special kind of mathematical program that can still be solved quite efficiently: a mixed linear program where some variables are required to be either 0 or 1. The adaptation and the resulting mixed-integer linear program is not straightforward. In fact, Koller, von Stegel and Megiddo could only find *one* Nash equilibrium in a 2-agent game. What is needed for DEC-POMDPs is to find the set of policies, called a joint policy, that corresponds to the Nash equilibrium with the highest value, finding “only” one Nash equilibrium – already a complex task – is not enough. Besides, whereas Koller, von Stegel and Megiddo algorithms could only be applied to 2-agent games, we extend the approach so as to solve DEC-POMDPs with an *arbitrary number of agents*, which constitutes an important contribution.

In order to formulate DEC-POMDPs as MILPs, we analyze in detail the structure of an optimal joint policy for a DEC-POMDP. A joint policy in sequence-form is expressed as a set of individual policies that are themselves described as a set of possible trajectories for each of the agents of the DEC-POMDP. Combinatorial considerations on these individual histories, as well as constraints that ensure these histories do define a valid joint policy are at the heart of the formulation of a DEC-POMDP as a mixed linear program, as developped in Sections 4 and 5. Thus, another contribution of our work is a better understanding of the properties of optimal solutions to DEC-POMDPs, a knowledge that might lead to the formulation of new approximate algorithms for DEC-POMDPs.

Another important contribution of this work is that we introduce **heuristics** for boosting the performance of the mathematical programs we propose (see Section 6). These heuristics take advantage of the succinctness of the DEC-POMDP model and of the knowledge acquired regarding the structure of optimal policies. Consequently, we are able to reduce the size of the mathematical programs (resulting also in reducing the time taken to solve them). These heuristics constitute an important pre-processing step in solving the programs. We present two types of heuristics: the elimination of *extraneous* histories which reduces the size of the mixed integer linear programs and the introduction of *cuts* in the mixed integer linear programs which reduces the time taken to solve a program.

On a more practical level, this article presents three different **mixed integer linear programs**, two are more directly derived from the work of Koller, von Stegel and Megiddo (see Table 4 and 5) and a third one is based solely on combinatorial considerations on the individual policies and histories (see Table 3). The theoretical validity of these formulations is backed by several theorems. We also conducted experimental evaluations of our algorithms and of our heuristics on several classical DEC-POMDP problems. We were thus able to confirm that our algorithms are quite comparable to dynamic programming exact algorithms but outperformed by forward search algorithms like GMAA* (Oliehoek et al., 2008). On some problems, though, MILPs are indeed faster by one order of magnitude or two than GMAA*.

### 1.4 Overview of this Article

The remainder of this article is organized as follows. Section 2 introduces the formalism of DEC-POMDP and some background on the classical algorithms, usually based on dynamic programing. Then we expose our reformulation of the DEC-POMDP in sequence-form in Section 3 where we also define various notions needed by the sequence-form. In Section 4, we show how to use combinatorial properties of the sequence-form policies to derive a first mixed integer linear program (MILP, in Table 3) for solving DEC-POMDP. By using game theoretic concepts like Nash equilibrium, we take inspiration from previous work on games in extensive form to design two other MILPs for solving DEC-POMDP (Tables 4, 5). These MILPs are smaller in size and their detailed derivation is presented in Section 5. Our contributed heuristics to speed up the practical resolutions of the various MILPs make up the core of Section 6. Section 7 presents experimental validations of our MILP-based algorithms on classical benchmarks of the DEC-POMDP literature as well as on randomly built problems. Finally, Section 8 analyzes and discusses our work and we conclude this paper with Section 9.

## 2. Dec-POMDP

This section gives a formal definition of Decentralized Partially Observed Markov Decision Processes as introduced by Bernstein et al. (2002). As described, a solution of a DEC-POMDP is a policy defined on the space of information sets that has an optimal value. This sections ends with a quick overview of the classical methods that have been developed to solve DEC-POMDPs.

### 2.1 Formal Definition

A **DEC-POMDP** is defined as a tuple $\mathcal{D} = \langle\, I,\, S,\, \{A_i\},\, \mathbb{P},\, \{O_i\},\, \mathbb{G},\, R,\, T,\, \alpha \,\rangle$ where:

- $I = \{1,\, 2,\, \cdots,\, n\}$ is a set of *agents*.
- $S$ is a finite set of *states*. The set of probability distributions over $S$ shall be denoted by $\Delta(S)$. Members of $\Delta(S)$ shall be called *belief states*.
- For each agent $i \in I$, $A_i$ is a set of *actions*. $A = \times_{i \in I} A_i$ denotes the set of joint actions.
- $\mathbb{P} : S \times A \times S \to [0,\, 1]$ is a *state transition function*. For each $s, s' \in S$ and for each $a \in A$, $\mathbb{P}(s, a, s')$ is the probability that the state of the problem in a period $t$ is $s'$ if, in period $t-1$, its state was $s$ and the agents performed the joint action $a$. Thus, for any time period $t \geq 2$, for each pair of states $s, s' \in S$ and for each joint action $a \in A$, there holds:

$$\mathbb{P}(s, a, s') \quad = \quad \Pr(s^t = s' | s^{t-1} = s, a^t = a).$$

  Thus, $(S,\, A,\, \mathbb{P})$ defines a discrete-state, discrete-time controlled Markov process.

- For each agent $i \in I$, $O_i$ is a set of *observations*. $O = \times_{i \in I} O_i$ denotes the set of joint observations.
- $\mathbb{G} : A \times S \times O \to [0,\, 1]$ is a *joint observation function*. For each $a \in A$, for each $o \in O$ and for each $s \in S$, $\mathbb{G}(a, s, o)$ is the probability that the agents receive the joint observation $o$ (that is, each agent $i$ receives the observation $o_i$) if the state of the problem in that period is $s$ and if in the previous period the agents took the joint action $a$. Thus, for any time period $t \geq 2$, for each joint action $a \in A$, for each state $s \in S$ and for each joint observation $o \in O$, there holds:

$$\mathbb{G}(a, s, o) \quad = \quad \Pr(o^t = o | s^t = s, a^{t-1} = a).$$

- $R : S \times A \to \mathbb{R}$ is a *reward function*. For each $s \in S$ and for each $a \in A$, $R(s, a) \in \mathbb{R}$ is the reward obtained by the agents if they take the joint action $a$ when the state of the process is $s$.

- $T$ is the *horizon* of the problem. The agents are allowed $T$ joint-actions before the process halts.
- $\alpha \in \Delta(S)$ is the *initial state* of the DEC-POMDP. For each $s \in S$, $\alpha(s)$ denotes the probability that the state of the problem in the first period is $s$.

As said, $S$, $A$ and $\mathbb{P}$ define a *controlled Markov Process* where the next state depends only on the previous state and on the joint action chosen by the agents. But the agents do not have access to the state of the process and can only rely on observations, generally partial and noisy, of this state, as specified by the observation function $\mathbb{G}$. From time to time, agents receive a non-zero reward according to the reward function $R$.

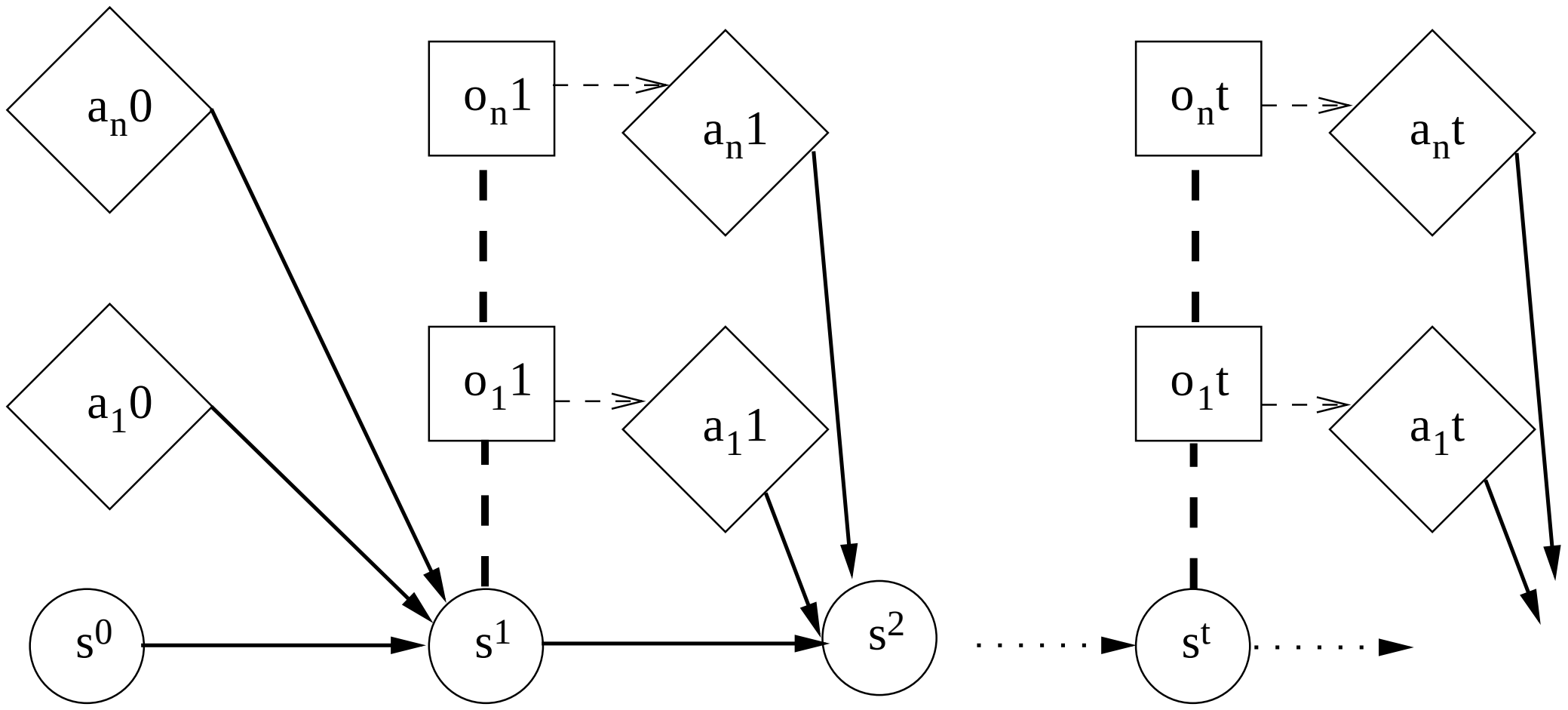

Figure 1: **DEC-POMDP**. At every period $t$ of the process, the environment is in state $s^t$, every agent $i$ receives observations $o_i^t$ and decides of its action $a_i^t$. The joint action $\langle a_1^t, a_2^t, \cdots, a_n^t\rangle$ alters the state of the process.

More specifically, as illustrated in Figure 1, the control of a DEC-POMDP by the $n$ agents unfolds over discrete time periods, $t = 1, 2, \cdots, T$ as follows. In each period $t$, the process is in a state denoted by $s^t$ from $S$. In the first period $t = 1$, the state $s^1$ is chosen according to $\alpha$ and the agents take actions $a_i^1$. In each period $t > 1$ afterward, each agent $i \in I$ takes an action denoted by $a_i^t$ from $A_i$ according to the agent's *policy*. When the agents take the joint action $a^t = \langle a_1^t, a_2^t, \cdots, a_n^t\rangle$, the following events occur:

1. The agents all obtain the same reward $R(s^t, a^t)$.
2. The state $s^{t+1}$ is determined according to the function $\mathbb{P}$ with arguments $s^t$ and $a^t$.
3. Each agent $i \in I$ receives an observation $o_i^{t+1}$ from $O_i$. The joint observation $o^{t+1} = \langle o_1^{t+1}, o_2^{t+1}, \cdots, o_n^{t+1}\rangle$ is determined by the function $\mathbb{G}$ with arguments $s^{t+1}$ and $a^t$.
4. The period changes from $t$ to $t + 1$.

In this paper, the DEC-POMDP we are interested in have the following properties:

- the horizon $T$ is finite and known by the agents;
- agents cannot infer the exact state of the system from their joint observations (this is the more general setting of DEC-POMDPs);
- agents do not observe actions and observations of the other agents. They are only aware of their own observations and reward;
- agents have a perfect memory of their past; they can base their choice of action on the sequence of past actions and observations. We speak of *perfect recall* setting;
- transition and observation functions are stationary, meaning that they do not depend on the period $t$.

Solving a DEC-POMDP means finding the agents' *policies* (*i.e., their decision functions*) to optimize a given criterion based on the rewards received. The criterion we will work with is called the **cumulative reward** and defined by:

$$\mathbb{E}\left[\sum_{t=1}^{T} R(s^t, \langle a_1^t, a_2^t, \dots, a_n^t\rangle)\right] \tag{1}$$

where $\mathbb{E}$ is the mathematical expectation.

## 2.2 Example of DEC-POMDP

The problem known as the "Decentralized Tiger Problem" (hereby denoted MA-Tiger), introduced by Nair, Tambe, Yokoo, Pynadath, and Marsella (2003), has been widely used to test DEC-POMDPs algorithms. It is a variation of a problem previously introduced for POMDPs (*i.e.*, DEC-POMDPs with one agent) by Kaelbling, Littman, and Cassandra (1998).

In this problem, we are given two agents confronted with two closed doors. Behind one door is a tiger, behind the other an escape route. The agents do not know which door leads to what. Each agent, independently of the other, can open one of the two doors or listen carefully in order to detect the tiger. If either of them opens the wrong door, the lives of both will be imperiled. If they both open the escape door, they will be free. The agents have a limited time in which to decide which door to open. They can use this time to gather information about the precise location of the tiger by listening carefully to detect the location of the tiger. This problem can be formalized as a DEC-POMDP with:

- two states as the tiger is either behind the left door ($s_l$) or the right door ($s_r$);
- two agents, that must decide and act;
- three actions for each agent: open the left door ($a_l$), open the right door ($a_r$) and listen ($a_o$);
- two observations, as the only thing the agent can observe is that they hear the tiger on the left ($o_l$) or on the right ($o_r$).

The initial state is chosen according to a uniform distribution over $S$. As long as the door remains closed, the state does not change but, when one door is opened, the state is reset to either $s_l$ or $s_r$ with equal probability. The observations are noisy, reflecting the difficulty of detecting the tiger. For example, when the tiger is on the left, the action $a_o$ produces an observation $o_l$ only 85% of the time. So if both agents perform $a_o$, the joint observation ($o_l$,$o_l$) occurs with a probability of $0.85 \times 0.85 = 0.72$. The reward function encourages the agents to coordinate their actions as, for example, the reward when both open the escape door (+20) is bigger than when one listens while the other opens the good door (+9). The full state transition function, joint observation function and reward function are described in the work of Nair et al. (2003).

### 2.3 Information Sets and Histories

An **information set** $\varphi$ of agent $i$ is a sequence $(a^1.o^2.a^2.o^3 \cdots .o^t)$ of even length in which the elements in odd positions are actions of the agent (members of $A_i$) and those in even positions are observations of the agent (members of $O_i$). An information set of length 0 shall be called the **null information set**, denoted by $\varnothing$. An information set of length $T-1$ shall be called a **terminal information set**. The set of information sets of lengths less than or equal to $T-1$ shall be denoted by $\Phi_i$.

We define a **history** of agent $i \in I$ to be a sequence $(a^1.o^2.\ a^2.\ o^3 \cdots .o^t.a^t)$ of odd length in which the elements in odd positions are actions of the agent (members of $A_i$) and those in even positions are observations of the agent (members of $O_i$). We define the **length** of a history to be the number of actions in the history ($t$ in our example). A history of length $T$ shall be called a **terminal history**. Histories of lengths less than $T$ shall be called **non-terminal** histories. The history of null length shall be denoted $\varnothing$. The information set associated to an history $h$, denoted $\varphi(h)$, is the information set composed by removing from $h$ its last action. If $h$ is a history and $o$ an observation, then $h.o$ is an information set.

We shall denote by $\mathcal{H}_i^t$ the set of all possible histories of length $t$ of agent $i$. Thus, $\mathcal{H}_i^1$ is just the set of actions $A_i$. We shall denote by $\mathcal{H}_i$ the set of histories of agent $i$ of lengths less than or equal to $T$. The size $n_i$ of $\mathcal{H}_i$ is thus:

$$n_i = |\mathcal{H}_i| \quad = \textstyle\sum_{t=1}^{T} |A_i|^t |O_i|^{t-1} \quad = |A_i| \frac{(|A_i||O_i|)^T - 1}{|A_i||O_i| - 1}. \tag{2}$$

The set $\mathcal{H}_i^T$ of terminal histories of agent $i$ shall be denoted by $\mathcal{E}_i$. The set $\mathcal{H}_i \backslash \mathcal{H}_i^T$ of non-terminal histories of agent $i$ shall be denoted by $\mathcal{N}_i$.

A tuple $\langle h_1, h_2, \ldots, h_n \rangle$ made of one history for each agent is called a **joint history**. The tuple obtained by removing the history $h_i$ from the joint history $h$ is noted $h_{-i}$ and called an $i$**-reduced joint history**.

**Example** Coming back to the MA-Tiger example, a set of valid histories could be: $\varnothing$, $(a_o)$, $(a_o.o_l.a_o)$, $(a_o.o_r.a_o)$, $(a_o.o_l.a_o.o_l.a_o)$, $(a_o.o_l.a_o.o_r.a_r)$, $(a_o.o_r.a_o.o_l.a_o)$ and $(a_o.o_r.a_o.o_r.a_r)$. Incidently, this set of histories corresponds to the support of the policy (*i.e.*, the histories generated by using this policy) of the Figure 2, as explained in the next section.

### 2.4 Policies

At each period of time, a policy must tell an agent what action to choose. This choice can be based on whatever past and present knowledge the agent has about the process at time $t$. One possibility is to define an **individual policy** $\pi_i$ of agent $i$ as a mapping from information sets to actions. More formally:

$$\pi_i \quad : \quad \Phi_i \longrightarrow \Delta(A_i) \tag{3}$$

Among the set $\Pi$ of policies, three families are usually distinguished:

- *Pure policies.* A pure or deterministic policy maps a given information set to *one* unique action. The set of pure policies for the agent $i$ is denoted $\hat{\Pi}$. Pure policies could also be defined using trajectories of past observations only since actions, which are chosen deterministically, can be reconstructed from the observations.
- *Mixed policies.* A mixed policy is a probability distribution over the set of pure policies. Thus, an agent using a mixed policy will control the DEC-POMDP by using a pure policy randomly chosen from a set of pure policies.
- *Stochastic policies.* A stochastic policy is the more general formulation as it associates a probability distribution over actions to each history.

If we come back to the MA-Tiger problem (Section 2.2), Figure 2 gives a possible policy for a horizon 2. As shown, a policy is classically represented by an action-observation tree. In that kind of tree, each branch is labelled by an observation. For a given sequence of past observations, one starts from the root node and follows the branches down to an action node. This node contains the action to be executed by the agent when it has seen this sequence of observations.

| Observation sequence | $\varnothing$ | $o_l$ | $o_r$ | $o_l.o_l$ | $o_l.o_r$ | $o_r.o_l$ | $o_r.o_r$ |
|---|---|---|---|---|---|---|---|
| Chosen action | $a_o$ | $a_o$ | $a_o$ | $a_l$ | $a_o$ | $a_o$ | $a_r$ |

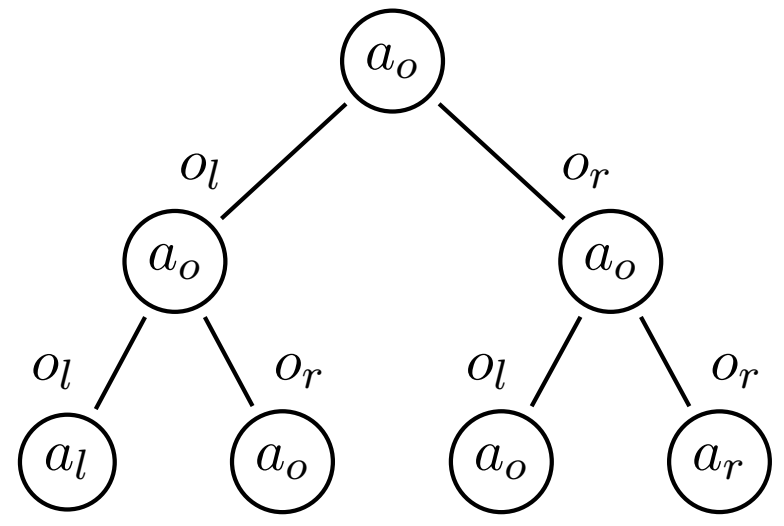

Figure 2: **Pure policy for MA-Tiger**. A pure policy maps sequences of observations to actions. This can be represented by an action-observation tree.

A **joint policy** $\pi = \langle \pi_1, \pi_2, \cdots, \pi_n \rangle$ is an $n$-tuple where each $\pi_i$ is a policy for agent $i$. Each of the individual policies must have the same horizon. For an agent $i$, we also define the notion of an ***i*-reduced joint policy** $\pi_{-i} = \langle \pi_1, \cdots, \pi_{i-1}, \pi_{i+1}, \cdots, \pi_n \rangle$ composed of the policies of all the other agents. We thus have that $\pi = \langle \pi_i, \pi_{-i} \rangle$.

### 2.5 Value Function

When executed by the agents, every $T$-horizon joint policy generates a probability distribution over the possible sequences of reward from which one can compute the **value** of the policy according to Equation 1. Thus the value of the joint policy $\pi$ is formally defined as:

$$V(\alpha, \pi) \quad = \quad \mathbb{E}\left[\sum_{t=1}^{T} R(s^t, a^t)|\pi, \alpha\right] \tag{4}$$

given that the state in the first period is chosen according to $\alpha$ and that actions are chosen according to $\pi$.

There is a recursive definition of the value function of a policy $\pi$ that is also a way to compute it when the horizon $T$ is finite. This definition requires some concepts that we shall now introduce.

Given a belief state $\beta \in \Delta(S)$, a joint action $a \in A$ and a joint observation $o \in O$, let $\mathcal{T}(o|\beta, a)$ denote the probability that the agents receive joint observation $o$ if they take joint action $a$ in a period $t$ in which the state is chosen according to $\beta$. This probability is defined as

$$\mathcal{T}(o|\beta, a) \quad = \quad \sum_{s \in S} \beta(s) \sum_{s' \in S} \mathbb{P}(s, a, s')\mathbb{G}(a, s', o) \tag{5}$$

Given a belief state $\beta \in \Delta(S)$, a joint action $a \in A$ and a joint observation $o \in O$ , the **updated belief state** $\beta^{ao} \in \Delta(S)$ of $\beta$ with respect to $a$ and $o$ is defined as (for each $s' \in S$),

$$\beta^{ao}(s') = \quad \frac{\mathbb{G}(a,s',o)\left[\sum_{s \in S} \beta(s)\mathbb{P}(s,a,s')\right]}{\mathcal{T}(o|\beta,a)} \quad \text{if } \mathcal{T}(o|\beta, a) > 0 \tag{6}$$

$$\beta^{ao}(s') = \quad 0 \quad \text{if } \mathcal{T}(o|\beta, a) = 0 \tag{7}$$

Given a belief state $\beta \in \Delta(S)$ and a joint action $a \in A$, $R(\beta, a)$ denotes $\sum_{s \in S} \beta(s)R(s, a)$. Using the above definitions and notations, the value $V(\alpha, \pi)$ of $\pi$ is defined as follows:

$$V(\alpha, \pi) \quad = \quad V(\alpha, \pi, \varnothing) \tag{8}$$

where $V(\alpha, \pi, \varnothing)$ is defined by recursion using equations (9), (10) and (11), given below. These equations are a straight reformulation of the classical Bellman equations for finite horizon problems.

- For histories of null length

$$V(\alpha, \pi, \varnothing) \quad = \quad R(\alpha, \pi(\varnothing)) + \sum_{o \in O} \mathcal{T}(o|\alpha, \pi(\varnothing))V(\alpha^{\pi(\varnothing)o}, \pi, o) \tag{9}$$

  $\pi(\varnothing)$ denotes the joint action $\langle \pi_1(\varnothing), \pi_2(\varnothing), \cdots, \pi_n(\varnothing)\rangle$ and $\alpha^{\pi(\varnothing)o}$ denotes the updated state of $\alpha$ given $\pi(\varnothing)$ and the joint observation $o$.

- For non-terminal histories. For any $\alpha' \in \Delta(S)$, for each $t$ of $\{1, \ldots, T-2\}$, for each tuple of sequences of $t$ observations $o^{1:T} = \langle o_1^{1:T},\, o_2^{1:T},\, \cdots,\, o_n^{1:T}\rangle$ where $o_i^{1:T} \in \times_T O_i$ is a sequence of $t$ observations of agent $i \in I$:

$$V(\alpha', \pi, o^{1:T}) \;=\; R(\alpha', \pi(o^{1:T})) + \sum_{o \in O} \mathcal{T}(o|\alpha', \pi(o^{1:T}))V(\alpha'^{\pi(o^{1:T})o}, \pi, o^{1:T}.o) \quad (10)$$

  $\alpha'^{\pi(o^{1:T})o}$ is the updated state of $\alpha'$ given the joint action $\pi(o^{1:T})$ and joint observation $o = \langle o_1, o_2, \cdots, o_n\rangle$ and $o^{1:T}.o$ is the tuple of sequences of $(t+1)$ observations $\langle o_1^{1:T}.o_1,\, o_2^{1:T}.o_2,\, \cdots,\, o_n^{1:T}.o_n\rangle$.

- For terminal histories. For any $\alpha' \in \Delta(S)$, for each tuple of sequences of ($T$ - 1) observations $o^{1:T-1} = \langle o_1^{1:T-1},\, o_2^{1:T-1},\, \cdots,\, o_n^{1:T-1}\rangle$:

$$V(\alpha', \pi, o^{1:T-1}) \;= R(\alpha, \pi(o^{1:T-1})) \;= \sum_{s \in S} \alpha'(s) R(s, \pi(o^{1:T-1})) \quad (11)$$

An **optimal policy** $\pi^*$ is a policy with the best possible value, verifying:

$$V(\alpha, \pi^*) \geq V(\alpha, \pi) \qquad \forall \pi \in \Pi. \quad (12)$$

An important fact about DEC-POMDPs, based on the following theorem, is that we can restrict ourselves to the set of pure policies when looking for a solution to a DEC-POMDP.

**Theorem 2.1.** *A DEC-POMDP has at least one optimal pure joint policy.*

*Proof:* See proof in the work of Nair et al. (2003). □

### 2.6 Overview of DEC-POMDPs Solutions and Limitations

As detailed in the work of Oliehoek et al. (2008), existing methods for solving DEC-POMDPs with finite-horizon belong to several broad families: "brute force", alternating maximization, search algorithms and dynamic programming.

**Brute Force** The simplest approach for solving a DEC-POMDP is to enumerate all possible joint policies and to evaluate them in order to find the optimal one. However, such a method becomes quickly *untractable* as the number of joint policies is doubly exponential in the horizon of the problem.

**Alternating Maximization** Following Chadès, Scherrer, and Charpillet (2002) and Nair et al. (2003), one possible way to solve DEC-POMDPs is for each agent (or each small group of agents) to alternatively search for a better policy while all the other agents freeze their own policy. Called *alternating maximization* by Oliehoek and *alternated co-evolution* by Chadès this method guarantees only to find a Nash equilibria, that is a *locally optimal joint policy.*

**Heuristic Search Algorithms** The concept was introduced by Szer, Charpillet, and Zilberstein (2005) and relies on heuristic search for looking for an optimal joint policy, using an admissible approximation of the value of the optimal joint policy. As the search progresses, joint policies that will provably by worse that the current admissible solution are pruned. Szer et al. used underlying MDPs or POMDPs to compute the admissible heuristic, Oliehoek et al. (2008) introduced a better heuristic based on the resolution of a Bayesian Game with a carefully crafted cost function. Currently, Oliehoek's method called GMAA* (for Generic Multi-Agent A*) is the quickest exact method on a large set of benchmarks. But, as every exact method, it is limited to quite simple problems.

**Dynamic Programming** The work from Hansen, Bernstein, and Zilberstein (2004) adapts solutions designed for POMDPs to the domain of DEC-POMDPs. The general idea is to start with policies for 1 time step that are used to build 2 time step policies and so on. But the process is clearly less efficient that the heuristic search approach as an exponential number of policies must be constructed and evaluated at each iteration of the algorithm. Some of these policies can be pruned but, once again, pruning is less efficient.

As exposed in more details in the paper by Oliehoek et al. (2008), several others approaches have been developed for subclasses of DEC-POMDPs. For example, special settings where agents are allowed to communicate and exchange informations or settings where the transition function can be split into independant transition functions for each agent have been studied and found easier to solve than "generic" DEC-POMDPs.

## 3. Sequence-Form of DEC-POMDPs

This section introduces the fundamental concept of policies in "sequence-form". A new formulation of a DEC-POMDP is thus possible and this leads to a Non-Linear Program (NLP) the solution of which defines an optimal solution to the DEC-POMDP.

### 3.1 Policies in Sequence-Form

A **history function** $p$ of an agent $i$ is a mapping from the set of histories to the interval $[0, 1]$. The value $p(h)$ is the **weight** of the history $h$ for the history function $p$. A policy $\pi_i$ defines a probability function over the set of histories of the agent $i$ by saying that, for each history $h_i$ of $\mathcal{H}_i$, $p(h_i)$ is the conditional probability of $h_i$ given an observation sequence $(o_i^0.o_i^1.\cdots.o_i^t)$ and $\pi_i$.

If every policy defines a policy function, not every policy function can be associated to a *valid* policy. Some constraints must be met. In fact, a history function $p$ is a **sequence-form policy** for agent $i$ when the following constraints are met:

$$\sum_{a \in A_i} p(a) = 1, \tag{13}$$

$$-p(h) + \sum_{a \in A_i} p(h.o.a) = 0, \quad \forall h \in \mathcal{N}_i, \forall o \in O_i, \tag{14}$$

where $h.o.a$ denotes the history obtained on concatenating $o$ and $a$ to $h$. This definition appears in a slightly different form as Lemma 5.1 in the work of Koller et al. (1994).

| |
|---|
| Variables: $x(h)$, $\forall h \in \mathcal{H}_i$, $$\sum_{a \in A_i} x(a) = 1 \quad (15)$$ $$-x(h) + \sum_{a \in A_i} x(h.o.a) = 0, \quad \forall h \in \mathcal{N}_i, \forall o \in O_i \quad (16)$$ $$x(h) \geq 0, \quad \forall h \in \mathcal{H}_i \quad (17)$$ |

Table 1: **Policy Constraints**. This set of linear inequalities, once solved, provide a valid sequence-form policy for the agent $i$. That is, from the weights $x(h)$, it is possible to define a policy for the agent $i$.

A sequence-form policy can be stochastic as the probability of choosing action $a$ in the information set $h.o$ is $p(h.o.a)/p(h)$. The **support** $S(p)$ of a sequence-form policy is made of the set of histories that have a non-negative weight, *i.e.* $S(p) = \{h \in \mathcal{H}_i \mid p(h) > 0\}$. As a sequence-form policy $p$ defines a unique policy $\pi$ for an agent, a sequence-form policy will be called a *policy* in the rest of this paper when no ambiguity is present.

The set of policies in the sequence-form of agent $i$ shall be denoted by $X_i$. The set of pure policies in the sequence-form shall be denoted by $\hat{X}_i \subset X_i$.

In a way similar to the definitions of Section 2.4, we define a **sequence-form joint policy** as a tuple of sequence-form policies, one for each agent. The weight of the joint history $h = \langle h_i \rangle$ of a sequence-form joint policy $\langle p_1, p_2, \cdots, p_n \rangle$ is the product $\prod_{i \in I} p_i(h_i)$. The set of joint policies in the sequence-form $\times_{i \in I} X_i$ shall be denoted by $X$ and the set of $i$-**reduced sequence-form joint policy** is called $X_{-i}$.

### 3.2 Policy Constraints

A policy of agent $i$ in the sequence-form can be found by solving a set of linear inequalities (LI) found in Table 1. These LI merely implement the definition of a policy in the sequence-form. The LI contains one variable $x(h)$ for each history $h \in \mathcal{H}_i$ to represent the weight of $h$ in the policy. A solution $x^*$ to these LI constitutes a policy in the sequence-form.

**Example** In the Section E.1 of the Appendices, the policy constraints for the decentralized Tiger problem are given for 2 agents and a horizon of 2.

Notice that in the policy constraints of an agent, each variable is only constrained to be non-negative whereas by the definition of a policy in sequence-form, the weight of a history must be in the interval $[0, 1]$. Does it mean that a variable in a solution to the policy constraints can assume a value higher than 1? Actually, the policy constraints are such that they prevent any variable from assuming a value higher than 1 as the following lemma shows.

**Lemma 3.1.** *In every solution $x^*$ to (15)-(17), for each $h \in \mathcal{H}_i$, $x^*(h)$ belongs to the $[0, 1]$ interval.*

*Proof:* This can be shown by forward induction.

Every $x(h)$ being non-negative (see Eq. (17)), it is also the case for every action $a$ of $A_i$. Then, no $x(a)$ can be greater than 1 otherwise constraint (15) would be violated. So, $\forall h \in \mathcal{H}_i^1$, (*i.e.* $\forall a \in A_i$), we have $x(h)$ belong to $[0, 1]$.

If every $h$ of $\mathcal{H}_i^t$ is such that $x(h) \in [0, 1]$, the previous reasoning applied using constraint (16) leads evidently to the fact that $x(h) \in [0, 1]$ for every $h$ of $\mathcal{H}_i^{t+1}$.

Thereby, by induction this holds for all $t$. □

Later in this article, in order to simplify the task of looking for joint policies, the policy constraints LI will be used to find pure policies. Looking for pure policies is not a limitation as finite-horizon DEC-POMDPs admit deterministic policies when the policies are defined on information set. In fact, pure policies are needed in two of the three MILPs we build in order to solve DEC-POMDPs, otherwise their derivation would not be possible (see Sections 4 and 5.4).

Looking for pure policies, an obvious solution would be to impose that every variable $x(h)$ belongs to the set $\{0, 1\}$. But, when solving a mixed integer linear program, it is generally a good idea to limit the number of integer variables as each integer variable is a possible node for the branch and bound method used to assign integer values to the variables. A more efficient implementation of a mixed integer linear program is to take advantage of the following lemma to impose that *only* the weights of the *terminal* histories take 0 or 1 as possible values.

**Lemma 3.2.** *If in* (15)-(17)*,* (17) *is replaced by,*

$$x(h) \geq 0, \quad \forall h \in \mathcal{N}_i \tag{18}$$

$$x(h) \in \{0, 1\}, \quad \forall h \in \mathcal{E}_i \tag{19}$$

*then in every solution $x^*$ to the resulting LI, for each $h \in \mathcal{H}_i$, $x^*(h) = 0$ or 1. We will speak of a 0-1 LI.*

*Proof:* We can prove this by backward induction. Let $h$ be a history of length $T$ - 1. Due to (16), for each $o \in O_i$, there holds,

$$x^*(h) = \sum_{a \in A_i} x^*(h.o.a). \tag{20}$$

Since $h$ is a history of length $T$ - 1, each history $h.o.a$ is a terminal history. Due to Lemma 3.1, $x^*(h) \in [0, 1]$. Therefore, the sum on the right hand side of the above equation is also in [0, 1]. But due to (19), each $x^*(h.o.a) \in \{0, 1\}$. Hence the sum on the right hand side is either 0 or 1, and not any value in between. Ergo, $x^*(h) \in \{0, 1\}$ and not any value in between. By this same reasoning, we can show that $x^*(h) \in \{0, 1\}$ for every non-terminal history $h$ of length $T$ - 2, $T$ - 3, $\cdots$, 1. □

To formulate the linear inequalities of Table 1 in memory, we require space that is only exponential in the horizon. For each agent $i \in I$, the size of $\mathcal{H}_i$ is $\sum_{t=1}^{T} |A_i|^t |O_i|^{t-1}$. It is then exponential in $T$ and the number of variables in the LP is also exponential in $T$. The number of constraints in the LI of Table 1 is $\sum_{t=0}^{T-1} |A_i|^t |O_i|^t$, meaning that the number of constraints of the LI is also exponential in $T$.

### 3.3 Sequence-Form of a DEC-POMDP

We are now able to give a formulation of a DEC-POMDP based on the use of sequence-form policies. We want to stress that this is *only* a re-formulation, but as such will provide us with new ways of solving DEC-POMDPs with mathematical programming.

Given a "classical" formulation of a DEC-POMDP (see Section 2.1), the equivalent **sequence-form DEC-POMDP** is a tuple $\langle I, \{\mathcal{H}_i\}, \Psi, \mathcal{R}\rangle$ where:

- $I = \{1, 2, \cdots, n\}$ is a set of agents.
- For each agent $i \in I$, $\mathcal{H}_i$ is the set of histories of length less than or equal to $T$ for the agent $i$, as defined in the previous section. Each set $\mathcal{H}_i$ is derived using the sets $A_i$ and $O_i$.
- $\Psi$ is the joint history conditional probability function. For each joint history $j \in \mathcal{H}$, $\Psi(\alpha, j)$ is the probability of $j$ occurring conditional on the agents taking joint actions according to it and given that the initial state of the DEC-POMDP is $\alpha$. This function is derived using the set of states $S$, the state transition function $\mathbb{P}$ and the joint observation function $\mathbb{G}$.
- $\mathcal{R}$ is the joint history value. For each joint history $j \in \mathcal{H}$, $\mathcal{R}(\alpha, j)$ is the value of the expected reward the agents obtain if the joint history $j$ occurs. This function is derived using the set of states $S$, the state transition function $\mathbb{P}$, the joint observation function $\mathbb{G}$ and the reward function $R$. Alternatively, $\mathcal{R}$ can be described as a function of $\Psi$ and $R$.

This formulation folds $S$, $\mathbb{P}$ and $\mathbb{G}$ into $\Psi$ and $\mathcal{R}$ by relying on the set of histories. We will now give more details about the computation of $\Psi$ and $\mathcal{R}$.

$\Psi(\alpha, j)$ is the **conditional probability** that the sequence of joint observations received by the agents till period $t$ is $(o^1(j).o^2(j).\cdots.\ o^{t-1}(j))$ **if** the sequence of joint actions taken by them till period $t$ - 1 is $(a^1(j).\ a^2(j).\ \cdots.\ a^{t-1}(j))$ **and** the initial state of the DEC-POMDP is $\alpha$. That is,

$$\Psi(\alpha, j) \quad = \quad \text{Prob.}(o^1(j).o^2(j).\cdots.o^{t-1}(j)|\alpha, a^1(j).a^2(j).\cdots.a^{t-1}(j)) \tag{21}$$

This probability is the product of the probabilities of seeing observation $o^k(j)$ given the appropriate belief state and action chosen at time $k$, that is:

$$\Psi(\alpha, j) \quad = \quad \prod_{k=1}^{t-1} \mathcal{T}(o^k(j)|\beta_j^{k-1}, a^k(j)) \tag{22}$$

where $\beta_j^k$ is the probability distribution on $S$ given that the agents have followed the joint history $j$ up to time $k$, that is:

$$\beta_j^k(s) \quad = \quad \text{Prob.}(s|o^1(j).a^1(j).\cdots.o^k(j)). \tag{23}$$

Variables: $x_i(h)$, $\forall i \in I$, $\forall h \in \mathcal{H}_i$

$$\text{Maximize} \quad \sum_{j \in \mathcal{E}} \mathcal{R}(\alpha, j) \prod_{i \in I} x_i(j_i) \tag{27}$$

subject to

$$\sum_{a \in A_i} x_i(a) \;=\; 1, \quad \forall i \in I \tag{28}$$

$$-x_i(h) + \sum_{a \in A_i} x_i(h.o.a) \;=\; 0, \quad \forall i \in I, \forall h \in \mathcal{N}_i, \forall o \in O_i \tag{29}$$

$$x_i(h) \;\geq\; 0, \quad \forall i \in I, \forall h \in \mathcal{H}_i \tag{30}$$

Table 2: **NLP**. This non-linear program expresses the constraints for finding a sequence-form joint policy that is an optimal solution to a DEC-POMDP.

Regarding the **value of a joint history**, it is defined by:

$$\mathcal{R}(\alpha, j) \;=\; \overline{R}(\alpha, j)\Psi(\alpha, j) \tag{24}$$

where

$$\overline{R}(\alpha, j) \;=\; \sum_{k=1}^{t} \sum_{s \in S} \beta_j^{k-1}(s) R(s, a^k(j)). \tag{25}$$

Thus, $\mathcal{V}(\alpha, p)$, the **value of a sequence-form joint policy** $p$, is the weighted sum of the value of the histories in its support:

$$\mathcal{V}(\alpha, p) \;=\; \sum_{j \in \mathcal{H}} p(j)\mathcal{R}(\alpha, j) \tag{26}$$

with $p(j) = \prod_{i \in I} p_i(j_i)$.

### 3.4 Non-Linear Program for Solving DEC-POMDPs.

By using the sequence-form formulation of a DEC-POMDP, we are able to express joint policies as sets of linear constraints and to assess the value of every policy. Solving a DEC-POMDP amounts to finding the policy with the maximal value, which can be done with the non-linear program (NLP) of Table 2 where, once again, the $x_i$ variables are the weights of the histories for the agent $i$.

**Example** An example of the formulation of such an NLP can be found in the Appendices, in Section E.2. It is given for the decentralized Tiger problem with 2 agents and an horizon of 2.

The constraints of the program form a convex set, but the objective function is not concave (as explained in appendix A). In the general case, solving non-linear program is very difficult and there are no generalized method that guarantee finding a global maximum point. However, this particular NLP is in fact a Multilinear Mathematical Program (see Drenick, 1992) and this kind of programs are still very difficult to solve. When only two agents are considered, one speaks of bilinear programs, that can be solved more easily (Petrik & Zilberstein, 2009; Horst & Tuy, 2003).

An evident, but inefficient, method to find a global maximum point is to evaluate *all* the extreme points of the set of feasible solutions of the program since it is known that every global as well as local maximum point of a non-concave function lies at an extreme point of such a set (Fletcher, 1987). This is an inefficient method because there is no test that tells if an extreme point is a local maximum point or a global maximum point. Hence, unless all extreme points are evaluated, we cannot be sure of having obtained a global maximum point. The set of feasible solutions to the NLP is $X$, the set of $T$-step joint policies. The set of extreme points of this set is $\hat{X}$, the set of pure $T$-step joint policies, whose number is doubly exponential in $T$ and exponential in $n$. So enumerating the extreme points for this NLP is untractable.

Our approach, developed in the next sections, is to linearize the objective function of this NLP in order to deal only with linear programs. We will describe two ways for doing this: one is based on combinatorial consideration (Section 4) and the other is based on game theory concepts (Section 5). In both cases, this shall mean adding more variables and constraints to the NLP, but upon doing so, we shall derive *mixed integer linear programs* for which it is possible to find a global maximum point and hence an optimal joint policy of the DEC-POMDP.

## 4. From Combinatorial Considerations to Mathematical Programming

This section explains how it is possible to use combinatorial properties of DEC-POMDPs to transform the previous NLP into a mixed integer linear program. As shown, this mathematical program belongs to the family of 0-1 Mixed Integer Linear Programs, meaning that some variables of this linear program must take integer values in the set $\{0, 1\}$.

### 4.1 Linearization of the Objective Function

Borrowing ideas from the field of *Quadratic Assignment Problems* (Papadimitriou & Steiglitz, 1982), we turn the non-linear objective function of the previous NLP into a linear objective function and linear constraints involving new variables $z$ that must take integer values. The variable $z(j)$ represents the product of the $x_i(j_i)$ variables.

Thus, the objective function that was:

$$\text{maximize} \quad \sum_{j \in \mathcal{E}} \mathcal{R}(\alpha, j) \prod_{i \in I} x_i(j_i) \tag{31}$$

can now be rewritten as

$$\text{maximize} \quad \sum_{j \in \mathcal{E}} \mathcal{R}(\alpha, j) z(j) \tag{32}$$

where $j = \langle j_1, j_2, \cdots, j_n \rangle$.

We must ensure that there is a two way mapping between the value of the new variables $z$ and the $x$ variables for any solution of the mathematical program, that is:

$$z^*(j) \quad = \quad \prod_{i \in I} x_i^*(j_i). \tag{33}$$

For this, we will restrict ourself to *pure policies* where the $x$ variables can only be 0 or 1. In that case, the previous constraint (33) becomes:

$$z^*(j) = 1 \quad \Leftrightarrow \quad x_i^*(j_i) = 1, \quad \forall i \in I \tag{34}$$

There, we take advantage on the fact that the support of a pure policy for an agent $i$ is composed of $|O_i|^{T-1}$ terminal histories to express these new constraints. On the one hand, to guarantee that $z(j)$ is equal to 1 only when enough $x$ variables are also equal to 1, we write:

$$\sum_{i=1}^{n} x_i(j_i) - nz(j) \quad \geq \quad 0, \quad \forall j \in \mathcal{E}. \tag{35}$$

On the other hand, to limit the number of $z(j)$ variables that can take a value of 1, we will enumerate the number of joint terminal histories to end up with:

$$\sum_{j \in \mathcal{E}} z(j) \quad = \quad \prod_{i \in I} |O_i|^{T-1}. \tag{36}$$

The constraints (35) would weight heavily on any mathematical program as there would be one constraint for each terminal *joint* history, a number which is exponential in $n$ and $T$. Our idea to reduce this number of constraints is not to reason about *joint histories* but with *individual histories*. An history $h$ of agent $i$ is part of the support of the solution of the problem (*i.e.*, $x_i(h) = 1$) if and only if the number of joint histories it belongs to ($\sum_{j' \in \mathcal{E}_{-i}} z(\langle h, j' \rangle)$) is $\prod_{k \in I \setminus \{i\}} |O_k|^{T-1}$. Then, we suggest to replace the $\prod |\mathcal{E}_i|$ constraints (35)

$$\sum_{i=1}^{n} x_i(j_i) - nz(j) \geq 0, \quad \forall j \in \mathcal{E}. \tag{35}$$

by the $\sum |\mathcal{E}_i|$ constraints

$$\begin{aligned} \sum_{j' \in \mathcal{E}_{-i}} z(\langle h, j' \rangle) \quad &= \quad \frac{\prod_{k \in I} |O_k|^{T-1}}{|O_i|^{T-1}} x_i(h) \\ &= \quad x_i(h) \prod_{k \in I \setminus \{i\}} |O_k|^{T-1}, \quad \forall i \in I, \forall h \in \mathcal{E}_i. \end{aligned} \tag{37}$$

### 4.2 Fewer Integer Variables

The linearization of the objective function rely on the fact that we are dealing with pure policies, meaning that every $x$ and $z$ variable is supposed to value either 0 or 1. As solving linear programs with integer variables is usually based on the "branch and bound" technique

Variables:
$x_i(h)$, $\forall i \in I$, $\forall h \in \mathcal{H}_i$
$z(j)$, $\forall j \in \mathcal{E}$

$$\text{Maximize} \qquad \sum_{j \in \mathcal{E}} \mathcal{R}(\alpha, j) z(j) \tag{38}$$

subject to:

$$\sum_{a \in A_i} x_i(a) = 1, \quad \forall i \in I \tag{39}$$

$$-x_i(h) + \sum_{a \in A_i} x_i(h.o.a) = 0, \quad \forall i \in I, \forall h \in \mathcal{N}_i, \forall o \in O_i \tag{40}$$

$$\sum_{j' \in H_{-i}^T} z(\langle h, j' \rangle) = x_i(h) \prod_{k \in I \setminus \{i\}} |O_k|^{T-1}, \quad \forall i \in I, \forall h \in \mathcal{E}_i \tag{41}$$

$$\sum_{j \in \mathcal{E}} z(j) = \prod_{i \in I} |O_i|^{T-1} \tag{42}$$

$$x_i(h) \geq 0, \quad \forall i \in I, \forall h \in \mathcal{N}_i \tag{43}$$

$$x_i(h) \in \{0, 1\}, \quad \forall i \in I, \forall h \in \mathcal{E}_i \tag{44}$$

$$z(j) \in [0, 1], \quad \forall j \in \mathcal{E} \tag{45}$$

Table 3: **MILP**. This 0-1 mixed integer linear program finds a sequence-form joint policy that is an optimal solution to a DEC-POMDP.

(Fletcher, 1987), for efficiency reasons, it is important to reduce the number of integer variables in our mathematical programs.

As done in Section 3.2, we can relax most $x$ variables and allow them to take non-negative values provided that the $x$ values for terminal histories are constrained to integer values. Furthermore, as proved by the following lemma, these constraints on $x$ *also* guarantee that $z$ variables only take their value in $\{0, 1\}$.

We eventually end up with the following linear program with real and integer variables, thus called an **0-1 mixed integer linear program** (MILP). The MILP is shown in Table 3.

**Example** In Section E.3 of the Appendices, an example if such MILP is given for the problem of the decentralized Tiger for 2 agents and an horizon of 2.

**Lemma 4.1.** *In every solution ($x^*$, $z^*$) to the* **MILP** *of Table 3, for each $j \in \mathcal{E}$, $z^*(j)$ is either 0 or 1.*

*Proof:* Let ($x^*$, $z^*$) be a solution of **MILP**. Let,

$$S(z) = \{j \in \mathcal{E} | z^*(j) > 0\} \tag{46}$$

$$S_i(x_i) = \{h \in \mathcal{E}_i | x_i^*(h) = 1\}, \quad \forall i \in I \tag{47}$$

$$S_i(z, j') = \{j \in \mathcal{E} | j_{-i} = j', z^*(j) > 0\}, \quad \forall i \in I, \forall j' \in \mathcal{E}_{-i} \tag{48}$$

Now, due to (42) and (45), $|S(z)| \geq \prod_{i \in I} |O_i|^{T-1}$. By showing that $|S(z)| \leq \prod_{i \in I} |O_i|^{T-1}$, we shall establish that $|S(z)| = \prod_{i \in I} |O_i|^{T-1}$. Then due to the upper bound of 1 on each $z$ variable, the implication will be that $z^*(j)$ is 0 or 1 for each terminal joint history $j$ thus proving the statement of the lemma.

Note that by Lemma (3.2), for each agent $i$, $x_i^*$ is a pure policy. Therefore, we have that $|S_i(x)| = |O_i|^{T-1}$. This means that in the set of constraints (41), an $i$-reduced terminal joint history $j' \in \mathcal{E}_{-i}$ will appear on the right hand side not more than $|O_i|^{T-1}$ times when in the left hand side, we have $x_i^*(h) = 1$. Thus, $\forall j' \in \mathcal{E}_{-i}$,

$$|S_i(z, j')| \quad \leq \quad |O_i|^{T-1}. \tag{49}$$

Now, we know that for each agent $i$ and for each history $h \in \mathcal{H}_i$ , $x_i^*(h)$ is either 0 or 1 since $x_i^*$ is a pure policy. So, given an $i$-reduced terminal joint history $j'$, $\prod_{k \in I \setminus \{i\}} x_k^*(j'_k)$ is either 0 or 1. Secondly, due to (41), the following implication clearly holds for each terminal joint history $j$,

$$z^*(j) > 0 \quad \Rightarrow \quad x_i^*(j_i) = 1, \quad \forall i \in I. \tag{50}$$

Therefore, we obtain

$$\begin{aligned} |S_i(z, j')| \quad &\leq \quad |O_i|^{T-1} &(51)\\ &= \quad |O_i|^{T-1} \prod_{k \in I \setminus \{i\}} x_k^*(j'_k). &(52) \end{aligned}$$

As a consequence,

$$\begin{aligned} \sum_{j' \in \mathcal{E}_{-i}} |S_i(z, j')| \quad &\leq \quad \sum_{j' \in \mathcal{E}_{-i}} |O_i|^{T-1} \prod_{k \in I \setminus \{i\}} x_k^*(j'_k) &(53)\\ &= \quad |O_i|^{T-1} \sum_{j' \in \mathcal{E}_{-i}} \prod_{k \in I \setminus \{i\}} x_k^*(j'_k) &(54)\\ &= \quad |O_i|^{T-1} \prod_{k \in I \setminus \{i\}} \sum_{h' \in \mathcal{E}_k} x_k^*(h') &(55)\\ &= \quad |O_i|^{T-1} \prod_{k \in I \setminus \{i\}} |O_k|^{T-1} &(56)\\ &= \quad \prod_{j \in I} |O_j|^{T-1}. &(57) \end{aligned}$$

Since $\bigcup_{j' \in \mathcal{E}_{-i}} S_i(z, j') = S(z)$, there holds that $\sum_{j' \in \mathcal{E}_{-i}} |S_i(z, j')| = |S(z)|$. Hence,

$$|S(z)| \quad \leq \quad \prod_{j \in I} |O_j|^{T-1}. \tag{58}$$

Thus the statement of the lemma is proved. □

### 4.3 Summary

By using combinatorial considerations, it is possible to design a 0-1 MILP for solving a given DEC-POMDP. As proved by theorem 4.1, the solution of this MILP defines an optimal joint policy for the DEC-POMDP. Nevertheless, this MILP is quite large, with $O(k^T)$ constraints and $\sum_i |\mathcal{H}_i| + \prod_i |\mathcal{E}_i| = O(k^{nT})$ variables, $O(k^T)$ of these variables must take integer values. The next section details another method for the linearization of NLP which leads to a "smaller" mathematical program for the 2-agent case.

**Theorem 4.1.** *Given a solution* $(x^*, z^*)$ *to* **MILP**, $x^* = \langle x_1^*, x_2^*, \cdots, x_n^* \rangle$ *is a pure* $T$*-period optimal joint policy in sequence-form.*

*Proof:* Due to the policy constraints and the domain constraints of each agent, each $x_i^*$ is a pure sequence-form policy of agent $i$. Due to the constraints (41)-(42), each $z^*$ values 1 if and only if the product $\prod_{i \in I} x_i(j_i)$ values 1. Then, by maximizing the objective function we are effectively maximizing the value of the sequence-form policy $\langle x_1^*, x_2^*, \cdots, x_n^* \rangle$. Thus, $\langle x_1^*, x_2^*, \cdots, x_n^* \rangle$ is an optimal joint policy of the original DEC-POMDP. □

## 5. From Game-Theoretical Considerations to Mathematical Programming

This section borrows concepts like "Nash equilibrium" and "regret" from game theory in order to design yet another 0-1 Mixed Integer Linear Program for solving DEC-POMDPs. In fact, two MILPs are designed, one that can only be applied for 2 agents and the other one for any number of agents. The main objective of this part is to derive a smaller mathematical program for the 2 agent case. Indeed, **MILP-2 agents** (see Table 4) has slightly less variables and constraints than **MILP** (see Table 3) and thus might prove easier to solve. On the other hand, when more than 2 agents are considered, the new derivation leads to a MILP which is only given for completeness as it is bigger than **MILP**.

Links between the fields of multiagent systems and game theory are numerous in the literature (see, for example, Sandholm, 1999; Parsons & Wooldridge, 2002). We will elaborate on the fact that the optimal policy of a DEC-POMDP is a Nash Equilibrium. It is in fact the Nash Equilibrium with the highest utility as the agents all share the same reward.

For the 2-agent case, the derivation we make in order to build the MILP is similar to the first derivation of Sandholm, Gilpin, and Conitzer (2005). We give more details of this derivation and adapt it to DEC-POMDP by adding an objective function to it. For more than 2 agents, our derivation can still be use to find Nash equilibriae with pure strategies.

For the rest of this article, we will make no distinction between a policy, a sequence-form policy or a strategy of an agent as, in our context, these concepts are equivalent. Borrowing from game theory, a joint policy will be denoted $p$ or $q$, an individual policy $p_i$ or $q_i$ and a $i$-reduced policy $p_{-i}$ or $q_{-i}$.

### 5.1 Nash Equilibrium

A Nash Equilibrium is a joint policy in which each policy is a best response to the reduced joint policy formed by the other policies of the joint policy. In the context of a sequence-form

DEC-POMDP, a policy $p_i \in X_i$ of agent $i$ is said to be a **best response** to an $i$-reduced joint policy $q_{-i} \in X_{-i}$ if there holds that

$$\mathcal{V}(\alpha, \langle p_i, q_{-i}\rangle) - \mathcal{V}(\alpha, \langle p'_i, q_{-i}\rangle) \quad \geq \quad 0, \quad \forall p'_i \in X_i. \tag{59}$$

A joint policy $p \in X$ is a **Nash Equilibrium** if there holds that

$$\mathcal{V}(\alpha, p) - \mathcal{V}(\alpha, \langle p'_i, p_{-i}\rangle) \quad \geq \quad 0, \quad \forall i \in I, \forall p'_i \in X_i. \tag{60}$$

That is,

$$\sum_{h\in\mathcal{E}_i}\sum_{j'\in\mathcal{E}_{-i}} \mathcal{R}(\alpha, \langle h, j'\rangle) \prod_{k\in I\backslash\{i\}} p_k(j'_k)\{p_i(h) - p'_i(h)\} \quad \geq \quad 0, \quad \forall i \in I, \forall p'_i \in X_i. \tag{61}$$

The derivation of the necessary conditions for a Nash equilibrium consists of deriving the necessary conditions for a policy to be a best response to a reduced joint policy. The following program finds a policy for an agent $i$ that is a best response to an $i$-reduced joint policy $q_{-i} \in X_{-i}$. Constraints (63)-(64) ensure that the policy defines a valid joint policy (see Section 3.2) and the objective function is a traduction of the concept of best response.

Variables: $x_i(h)$, $\forall i \in I$, $\forall h \in \mathcal{H}_i$

$$\text{Maximize} \quad \sum_{h\in\mathcal{E}_i}\left\{\sum_{j'\in\mathcal{E}_{-i}} \mathcal{R}(\alpha, \langle h, j'\rangle) \prod_{k\in I\backslash\{i\}} q_k(j'_k)\right\} x_i(h) \tag{62}$$

subject to:

$$\sum_{a\in A_i} x_i(a) \quad = \quad 1 \tag{63}$$

$$-x_i(h) + \sum_{a\in A_i} x_i(h.o.a) \quad = \quad 0, \quad \forall h \in \mathcal{N}_i, \forall o \in O_i \tag{64}$$

$$x_i(h) \quad \geq \quad 0, \quad \forall h \in \mathcal{H}_i. \tag{65}$$

This linear program (LP) must still be refined so that its solution is not only a best response for agent $i$ but a “global” best response, *i.e.*, the policy of *each* agent is a best response to all the other agents. This will mean introducing new variables (a set of variable for each agent). The main point will be to adapt the objective function as the current objective function, when applied to find “global” best response, would lead to a non-linear objective function where product of weights of policies would appear. To do this, we will make use of the *dual* of the program (LP).

The linear program (LP) has one variable $x_i(h)$ for each history $h \in \mathcal{H}_i$ representing the weight of $h$. It has one constraint per information set of agent $i$. In other words, each constraint of the linear program (LP) is uniquely labeled by an information set. For instance, the constraint (63) is labeled by the null information set $\varnothing$, and for each nonterminal history $h$ and for each observation $o$, the corresponding constraint in (64) is labeled by the information set $h.o$. Thus, (LP) has $n_i$ variables and $m_i$ constraints.

As described in the appendix (see appendix B), the dual of (LP) is expressed as:

Variables: $y_i(\varphi)$, $\forall \varphi \in \Phi_i$

$$\text{Minimize} \quad y_i(\varnothing) \tag{66}$$

subject to:

$$y_i(\varphi(h)) - \sum_{o \in O_i} y_i(h.o) \quad \geq \quad 0, \quad \forall h \in \mathcal{N}_i \tag{67}$$

$$y_i(\varphi(h)) - \sum_{j' \in \mathcal{E}_{-i}} \mathcal{R}(\alpha, \langle h, j' \rangle) \prod_{k \in I \backslash \{i\}} q_k(j'_k) \quad \geq \quad 0, \quad \forall h \in \mathcal{E}_i \tag{68}$$

$$y_i(\varphi) \quad \in \quad (-\infty, +\infty), \quad \forall \varphi \in \Phi_i \tag{69}$$

where $\varphi(h)$ denotes the information set to which $h$ belongs. The dual has one free variable $y_i(\dot{})$ for every *information set* of agent $i$. This is why the function $\varphi(h)$ (defined in Section 2.3) appears as a mapping from histories to information sets[1]. The dual program has one constraint per history of the agent. Thus, the dual has $m_i$ variables and $n_i$ constraints. Note that the objective of the dual is to minimize only $y_i(\varnothing)$ because in the primal (LP), the right hand side of all the constraints, except the very first one, is a 0.

The theorem of duality (see the appendix B), applied to the primal (LP) (62)-(65) and the transformed dual (66)-(69), says that their solutions have the same value. Mathematically, that means that:

$$\sum_{h \in \mathcal{E}_i} \left\{ \sum_{j' \in \mathcal{E}_{-i}} \mathcal{R}(\alpha, \langle h, j' \rangle) \prod_{k \in I \backslash \{i\}} q_k(j'_k) \right\} x_i^*(h) \quad = \quad y_i^*(\varnothing). \tag{70}$$

Thus, the value of the joint policy $\langle x_i^*, q_{-i} \rangle$ can be expressed either as

$$\mathcal{V}(\alpha, \langle x_i^*, q_{-i} \rangle) \quad = \quad \sum_{h \in \mathcal{E}_i} \{ \sum_{j' \in \mathcal{E}_{-i}} \mathcal{R}(\alpha, \langle h, j' \rangle) \prod_{k \in I \backslash \{i\}} q_k(j'_k) \} x_i^*(h) \tag{71}$$

or as

$$\mathcal{V}(\alpha, \langle x_i^*, q_{-i} \rangle) \quad = \quad y_i^*(\varnothing). \tag{72}$$

Due to the constraints (63) and (64) of the primal LP, there holds that

$$y_i^*(\varnothing) \quad = \quad y_i^*(\varnothing) \{ \sum_{a \in A_i} x_i^*(a) \} + \sum_{h \in \mathcal{N}_i} \sum_{o \in O_i} y_i^*(h.o) \{ -x_i^*(h) + \sum_{a \in A_i} x_i^*(h.o.a) \} \tag{73}$$

as constraint (63) guarantees that the first term in the braces is 1 and constraints (65) guarantee that each of the remaining terms inside the braces is 0. The right hand side of (73) can be rewritten as

$$\begin{aligned} \sum_{a \in A_i} x_i^*(a) \{ y_i^*(\varnothing) - \textstyle\sum_{o \in O_i} y_i^*(a.o) \} & + \sum_{h \in \mathcal{N}_i \backslash A_i} x_i^*(h) \{ y_i^*(\varphi(h)) - \sum_{o \in O_i} y_i^*(h.o) \} \\ & + \sum_{h \in \mathcal{E}_i} x_i^*(h) y_i^*(\varphi(h)) \\ = \textstyle\sum_{h \in \mathcal{N}_i} x_i^*(h) \{ y_i^*(\varphi(h)) - & \sum_{o \in O_i} y_i^*(h.o) \} + \sum_{h \in \mathcal{E}_i} x_i^*(h) y_i^*(\varphi(h)) \end{aligned} \tag{74}$$

1. As $h.o$ is an information set, $y_i(h.o)$ is a shortcut in writing for $y_i(\varphi(h.o))$.

So, combining equations (70) and (74), we get

$$\sum_{h\in\mathcal{N}_i} x_i^*(h) \qquad \left\{y_i^*(\varphi(h)) - \sum_{o\in O_i} y_i^*(h.o)\right\} + \sum_{h\in\mathcal{E}_i} x_i^*(h)\left\{y_i^*(\varphi(h)) - \sum_{j'\in\mathcal{E}_{-i}} \mathcal{R}(\alpha,\langle h,j'\rangle) \prod_{k\in I\setminus\{i\}} q_k(j'_k)\right\} = 0 \qquad (75)$$

It is time to introduce supplementary variables $w$ for each information set. These variables, usually called *slack* variables, are defined as:

$$y_i(\varphi(h)) - \sum_{o\in O_i} y_i(h.o) \;=\; w_i(h), \quad \forall h\in\mathcal{N}_i \qquad (76)$$

$$y_i(\varphi(h)) - \sum_{j'\in\mathcal{E}_{-i}} \mathcal{R}(\alpha,\langle h,j'\rangle) \prod_{k\in I\setminus\{i\}} q_k(j'_k) \;=\; w_i(h), \quad \forall h\in\mathcal{E}_i. \qquad (77)$$

As shown is Section C of the appendix, these slack variables correspond to the concept of regret as defined in game theory. The regret of an history expresses the loss in accumulated reward the agent incurs when he acts according to this history rather than according to a history which would belong to the optimal joint policy.

Thanks to the slack variables, we can furthermore rewrite (75) as simply

$$\sum_{h\in\mathcal{N}_i} x_i^*(h)w_i^*(h) + \sum_{h\in\mathcal{E}_i} x_i^*(h)w_i^*(h) \;=\; 0 \qquad (78)$$

$$\sum_{h\in\mathcal{H}_i} x_i^*(h)w_i^*(h) \;=\; 0. \qquad (79)$$

Now, (79) is a sum of $n_i$ products, $n_i$ being the size of $\mathcal{H}_i$. Each product in this sum is necessarily 0 because both $x_i(h)$ and $w_i(h)$ are constrained to be nonnegative in the primal and the dual respectively. This property is strongly linked to the complementary slackness optimality criterion in linear programs (see, for example, Vanderbei, 2008). Hence, (79) is equivalent to

$$x_i^*(h)w_i^*(h) \;=\; 0, \quad \forall h\in\mathcal{H}_i. \qquad (80)$$

Back to the framework of DEC-POMDPs, these constraints are written:

$$p_i(h)\mu_i(\langle h,q_{-i}\rangle) \;=\; 0, \quad \forall h\in\mathcal{H}_i. \qquad (81)$$

To sum up, solving the following mathematical program would give an optimal joint policy for the DEC-POMDP. But constraints (87) are non-linear and thus prevent us from solving this program directly. The linearization of these constraints, called complementarity constraints, is the subject of the next section.

Variables:
$x_i(h)$, $w_i(h)$ $\forall i\in I$ and $\forall h\in\mathcal{H}_i$
$y_i(\varphi)$ $\forall i\in I$ and $\forall\varphi\in\Phi_i$

$$\text{Maximize} \quad y_1(\varnothing) \qquad (82)$$

subject to:

$$\sum_{a \in A_i} x_i(a) = 1 \quad (83)$$

$$-x_i(h) + \sum_{a \in A_i} x_i(h.o.a) = 0, \quad \forall i \in I, \forall h \in \mathcal{N}_i, \forall o \in O_i \quad (84)$$

$$y_i(\varphi(h)) - \sum_{o \in O_i} y_i(h.o) = w_i(h), \quad \forall i \in I, \forall h \in \mathcal{N}_i \quad (85)$$

$$y_i(\varphi(h)) - \sum_{j' \in \mathcal{E}_{-i}} \mathcal{R}(\alpha, \langle h, j' \rangle) \prod_{k \in I \backslash \{i\}} x_k(j'_k) = w_i(h), \quad \forall i \in I, \forall h \in \mathcal{E}_i \quad (86)$$

$$x_i(h) w_i(h) = 0, \quad \forall i \in I, \forall h \in \mathcal{H}_i \quad (87)$$

$$x_i(h) \geq 0, \quad \forall i \in I, \forall h \in \mathcal{H}_i \quad (88)$$

$$w_i(h) \geq 0, \quad \forall i \in I, \forall h \in \mathcal{H}_i \quad (89)$$

$$y_i(\varphi) \in (-\infty, +\infty), \quad \forall i \in I, \forall \varphi \in \Phi_i \quad (90)$$

### 5.2 Dealing with Complementarity Constraints

This section explains how the non-linear constraints $x_i(h)w_i(h) = 0$ in the previous mathematical program can be turned into sets of linear constraints and thus lead to a mixed integer linear programming formulation of the solution of a DEC-POMDP.

Consider a complementarity constraint $ab = 0$ in variables $a$ and $b$. Assume that the lower bound on the values of $a$ and $b$ is 0. Let the upper bounds on the values of $a$ and $b$ be respectively $u_a$ and $u_b$. Now let $c$ be a 0-1 variable. Then, the complementarity constraint $ab = 0$ can be separated into the following equivalent pair of linear constraints,

$$a \leq u_a c \quad (91)$$

$$b \leq u_b(1 - c). \quad (92)$$

In other words, if this pair of constraints is satisfied, then it is surely the case that $ab = 0$. This is easily verified. $c$ can either be 0 or 1. If $c = 0$, then $a$ will be set to 0 because $a$ is constrained to be no more than $u_a c$ (and not less than 0); if $c = 1$, then $b$ will be set to 0 since $b$ is constrained to be not more than $u_b(1 - c)$ (and not less than 0). In either case, $ab = 0$.

Now consider each complementarity constraint $x_i(h)w_i(h) = 0$ from the non-linear program (82)-(90) above. We wish to separate each constraint into a pair of linear constraints. We recall that $x_i(h)$ represents the weight of $h$ and $w_i(h)$ represents the regret of $h$. The first requirement to convert this constraint to a pair of linear constraints is that the lower bound on the values of the two terms be 0. This is indeed the case since $x_i(h)$ and $w_i(h)$ are both constrained to be non-negative in the NLP. Next, we require upper bounds on the weights of histories and regrets of histories. We have shown in Lemma 3.1 that the upper bound on the value of $x_i(h)$ for each $h$ is 1. For the upper bounds on the regrets of histories, we require some calculus.

In any policy $p_i$ of agent $i$ there holds that

$$\sum_{h \in \mathcal{E}_i} p_i(h) \quad = \quad |O_i|^{T-1}. \tag{93}$$

Therefore, in every $i$-reduced joint policy $\langle q_1, q_2, \cdots, q_n \rangle \in X_{-i}$, there holds

$$\sum_{j' \in \mathcal{E}_{-i}} \prod_{k \in I \backslash \{i\}} q_k(j'_k) \quad = \quad \prod_{k \in I \backslash \{i\}} |O_k|^{T-1} \tag{94}$$

Since the regret of a terminal history $h$ of agent $i$ given $\langle q_1, q_2, \cdots, q_n \rangle$ is defined as

$$\mu_i(h, q) \quad = \quad \max_{h' \in \varphi(h)} \sum_{j' \in \mathcal{E}_{-i}} \prod_{k \in I \backslash \{i\}} q_k(j'_k) \left\{ \mathcal{R}(\alpha, \langle h', j' \rangle) - \mathcal{R}(\alpha, \langle h, j' \rangle) \right\}, \tag{95}$$

we can conclude that an **upper bound** $\mathcal{U}_i(h)$ on the regret of a **terminal history** $h \in \mathcal{E}_i$ of agent $i$ is,

$$\mathcal{U}_i(h) \quad = \quad \prod_{k \in I \backslash \{i\}} |O_k|^{T-1} \left\{ \max_{h' \in \varphi(h)} \max_{j' \in \mathcal{E}_{-i}} \mathcal{R}(\alpha, \langle h', j' \rangle) - \min_{j'' \in \mathcal{E}_{-i}} \mathcal{R}(\alpha, \langle h, j'' \rangle) \right\}. \tag{96}$$

Now let us consider the upper bounds on the regrets of non-terminal histories. Let $\varphi$ be an information set of length $t$ of agent $i$. Let $\mathcal{E}_i(\varphi) \subseteq \mathcal{E}_i$ denote the set of terminal histories of agent $i$ such the first $2t$ elements of each history in the set are identical to $\varphi$. Let $h$ be a history of length $t \leq T$ of agent $i$. Let $\mathcal{E}_i(h) \subseteq \mathcal{E}_i$ denote the set of terminal histories such that the first $2t$ - 1 elements of each history in the set are identical to $h$. Since in any policy $p_i$ of agent $i$, there holds

$$\sum_{h' \in \mathcal{E}_i(h)} p_i(h') \leq |O_i|^{T-t} \tag{97}$$

we can conclude that an **upper bound** $\mathcal{U}_i(h)$ on the regret of a **nonterminal history** $h \in \mathcal{N}_i$ of length $t$ agent $i$ is

$$\mathcal{U}_i(h) = L_i \left\{ \max_{h' \in \mathcal{E}_i(\varphi(h))} \max_{j' \in \mathcal{E}_{-i}} \mathcal{R}(\alpha, \langle h', j' \rangle) - \min_{g \in \mathcal{E}_i(h)} \min_{j'' \in \mathcal{E}_{-i}} \mathcal{R}(\alpha, \langle g, j'' \rangle) \right\} \tag{98}$$

where

$$L_i \quad = \quad |O_i|^{T-t} \prod_{k \in I \backslash \{i\}} |O_k|^{T-1}. \tag{99}$$

Notice that if $t = T$ (that is, $h$ is terminal) (98) reduces to (96).

So, the complementarity constraint $x_i(h) w_i(h) = 0$ can be separated into a pair of linear constraints by using a 0-1 variable $b_i(h)$ as follows,

$$x_i(h) \quad \leq \quad 1 - b_i(h) \tag{100}$$

$$w_i \quad \leq \quad \mathcal{U}_i(h) b_i(h) \tag{101}$$

$$b_i(h) \quad \in \quad \{0, 1\} \tag{102}$$

Variables:
$x_i(h)$, $w_i(h)$ and $b_i(h)$ for $i \in \{1,2\}$ and $\forall h \in \mathcal{H}_i$
$y_i(\varphi)$ for $i \in \{1,2\}$ and $\forall \varphi \in \Phi_i$

$$\text{Maximize} \quad y_1(\varnothing) \tag{103}$$

subject to:

$$\sum_{a \in A_i} x_i(a) = 1 \tag{104}$$

$$-x_i(h) + \sum_{a \in A_i} x_i(h.o.a) = 0, \quad i = 1, 2, \forall h \in \mathcal{N}_i, \forall o \in O_i \tag{105}$$

$$y_i(\varphi(h)) - \sum_{o \in O_i} y_i(h.o) = w_i(h), \quad i = 1, 2, \forall h \in \mathcal{N}_i \tag{106}$$

$$y_1(\varphi(h)) - \sum_{h' \in \mathcal{E}_2} \mathcal{R}(\alpha, \langle h, h' \rangle) x_2(h') = w_1(h), \quad \forall h \in \mathcal{E}_1 \tag{107}$$

$$y_2(\varphi(h)) - \sum_{h' \in \mathcal{E}_1} \mathcal{R}(\alpha, \langle h', h \rangle) x_1(h') = w_2(h), \quad \forall h \in \mathcal{E}_2 \tag{108}$$

$$x_i(h) \leq 1 - b_i(h), \quad i = 1, 2, \forall h \in \mathcal{H}_i \tag{109}$$

$$w_i(h) \leq \mathcal{U}_i(h) b_i(h), \quad i = 1, 2, \forall h \in \mathcal{H}_i \tag{110}$$

$$x_i(h) \geq 0, \quad i = 1, 2, \forall h \in \mathcal{H}_i \tag{111}$$

$$w_i(h) \geq 0, \quad i = 1, 2, \forall h \in \mathcal{H}_i \tag{112}$$

$$b_i(h) \in \{0, 1\}, \quad i = 1, 2, \forall h \in \mathcal{H}_i \tag{113}$$

$$y_i(\varphi) \in (-\infty, +\infty), \quad i = 1, 2, \forall \varphi \in \Phi_i \tag{114}$$

Table 4: **MILP-2 agents**. This 0-1 mixed integer linear program, derived from game theoretic considerations, finds optimal *stochastic* joint policies for DEC-POMDPs with 2 agents.

### 5.3 Program for 2 Agents

When we combine the policy constraints (Section 3.2), the constraints we have just seen for a policy to be a best response (Sections 5.1, 5.2) and a maximization of the value of the joint policy, we can derive a 0-1 mixed integer linear program the solution of which is an optimal joint policy for a DEC-POMDP for 2 agents. Table 4 details this program that we will call **MILP-2 agents**.

**Example** The formulation of the decentralized Tiger problem for 2 agents and for an horizon of 2 can be found in the appendices, in Section E.4

The variables of the program are the vectors $x_i$, $w_i$, $b_i$ and $y_i$ for each agent $i$. Note that for each agent $i \in I$ and for each history $h$ of agent $i$, $\mathcal{U}_i(h)$ denotes the upper bound on the regret of history $h$.

A solution ($x^*$, $y^*$, $w^*$, $b^*$) to **MILP-2 agents** consists of the following quantities: (i) an optimal joint policy $x^* = \langle x_1^*, x_2^* \rangle$ which may be stochastic; (ii) for each agent $i = 1$, 2, for each history $h \in \mathcal{H}_i$, $w_i^*(h)$, the regret of $h$ given the policy $x_{-i}^*$ of the other agent; (iii) for each agent $i = 1$, 2, for each information set $\varphi \in \Phi_i$, $y_i^*(\varphi)$, the value of $\varphi$ given the policy $x_{-i}$ of the other agent; (iv) for each agent $i = 1$, 2, the vector $b_i^*$ simply tells us which histories are not in the support of $x_i^*$; each history $h$ of agent $i$ such that $b_i^*(h) = 1$ is *not* in the support of $x_i^*$. Note that we can replace $y_1(\varnothing)$ by $y_2(\varnothing)$ in the objective function without affecting the program. We have the following result.

**Theorem 5.1.** *Given a solution ($x^*$, $w^*$, $y^*$, $b^*$) to* **MILP-2 agents**, $x^* = \langle x_1^*, x_2^* \rangle$ *is an optimal joint policy in sequence-form.*

*Proof:* Due to the policy constraints of each agent, each $x_i^*$ is a sequence-form policy of agent $i$. Due to the constraints (106)-(108), $y_i^*$ contains the values of the information sets of agent $i$ given $x_{-i}^*$. Due to the complementarity constraints (109)-(110), each $x_i^*$ is a best response to $x_{-i}^*$. Thus $\langle x_1^*, x_2^* \rangle$ is a Nash equilibrium. Finally, by maximizing the value of the null information set of agent 1, we are effectively maximizing the value of $\langle x_1^*, x_2^* \rangle$. Thus $\langle x_1^*, x_2^* \rangle$ is an optimal joint policy. □

In comparison with the **MILP** presented before in Table 3, **MILP-2 agents** should constitutes a particularly effective program in term of computation time for finding a 2-agent optimal $T$-period joint policy because it is a much smaller program. While the number of variables required by **MILP** is exponential in $T$ and in $n$, the number of variables required by **MILP-2 agents** is exponential only in $T$. This represents a major reduction in size that should lead to an improvement in term of computation time.

### 5.4 Program for 3 or More Agents

When the number of agents is more than 2, the constraint (86) of the non-linear program (82)-(90) is no longer a complementarity constraint between 2 variables that could be linearized as before. In particular, the term $\prod_{k \in I \setminus \{i\}} x_k(j'_k)$ of the constraint (86) involves as many variables as there are different agents. To linearize this term, we will restrict ourselves once again to *pure* joint policies and exploit some combinatorial facts on the number of histories involved. This leads to the 0-1 mixed linear program called **MILP-$n$ agents** and depicted in Table 5.

The variables of the program **MILP-$n$ agents** are the vectors $x_i$, $w_i$, $b_i$ and $y_i$ for each agent $i$ and the vector $z$. We have the following result.

**Theorem 5.2.** *Given a solution ($x^*$, $w^*$, $y^*$, $b^*$, $z^*$) to* **MILP-$n$ agents**, $x^* = \langle x_1^*, x_2^*, \cdots, x_n^* \rangle$ *is a pure $T$-period optimal joint policy in sequence-form.*

*Proof:* Due to the policy constraints and the domain constraints of each agent, each $x_i^*$ is a pure sequence-form policy of agent $i$. Due to the constraints (118)-(119), each $y_i^*$ contains the values of the information sets of agent $i$ given $x_{-i}^*$. Due to the complementarity constraints (122)-(123), each $x_i^*$ is a best response to $x_{-i}^*$. Thus $x^*$ is a Nash equilibrium. Finally, by maximizing the value of the null information set of agent 1, we are effectively maximizing the value of $x^*$. Thus $x^*$ is an optimal joint policy. □

Variables:
$x_i(h)$, $w_i(h)$ and $b_i(h)$ $\forall i \in I$ and $\forall h \in \mathcal{H}_i$
$y_i(\varphi)$ $\forall i \in I$, $\forall \varphi \in \Phi_i$
$z(j)$ $\forall j \in \mathcal{E}$

$$\text{Maximize} \quad y_1(\varnothing) \tag{115}$$

subject to:

$$\sum_{a \in A_i} x_i(a) = 1 \tag{116}$$

$$-x_i(h) + \sum_{a \in A_i} x_i(h.o.a) = 0, \quad \forall i \in I, \forall h \in \mathcal{N}_i, \forall o \in O_i \tag{117}$$

$$y_i(\varphi(h)) - \sum_{o \in O_i} y_i(h.o) = w_i(h), \quad \forall i \in I, \forall h \in \mathcal{N}_i \tag{118}$$

$$y_i(\varphi(h)) - \frac{1}{|O_i|^{T-1}} \sum_{j \in \mathcal{E}} \mathcal{R}(\alpha, \langle h, j_{-i} \rangle) z(j) = w_i(h), \quad \forall i \in I, \forall h \in \mathcal{E}_i \tag{119}$$

$$\sum_{j' \in \mathcal{E}_{-i}} z(\langle h, j' \rangle) = x_i(h) \prod_{k \in I \setminus \{i\}} |O_k|^{T-1}, \quad \forall i \in I, \forall h \in \mathcal{E}_i \tag{120}$$

$$\sum_{j \in \mathcal{E}} z(j) = \prod_{i \in I} |O_i|^{T-1} \tag{121}$$

$$x_i(h) \leq 1 - b_i(h), \quad \forall i \in I, \forall h \in \mathcal{H}_i \tag{122}$$

$$w_i(h) \leq \mathcal{U}_i(h) b_i(h), \quad \forall i \in I, \forall h \in \mathcal{H}_i \tag{123}$$

$$x_i(h) \geq 0, \quad \forall i \in I, \forall h \in \mathcal{N}_i \tag{124}$$

$$x_i(h) \in \{0,1\} \quad \forall i \in I, \forall h \in \mathcal{E}_i \tag{125}$$

$$w_i(h) \geq 0, \quad \forall i \in I, \forall h \in \mathcal{H}_i \tag{126}$$

$$b_i(h) \in \{0,1\}, \quad \forall h \in \mathcal{H}_i \tag{127}$$

$$y_i(\varphi) \in (-\infty, +\infty), \quad \forall i \in I, \forall \varphi \in \Phi_i \tag{128}$$

$$z(j) \in [0,1], \quad \forall j \in \mathcal{E} \tag{129}$$

Table 5: **MILP-$n$ agents**. This 0-1 mixed integer linear program, derived from game theoretic considerations, finds *pure* optimal joint policies for DEC-POMDPs with 3 or more agents.

Compared to the **MILP** of Table 3, **MILP-$n$ agents** has roughly the same size but with more real valued variables and more 0-1 variables. To be precise, **MILP** has a 0-1 variable for every terminal history of every agent (that is *approximatively* $\sum_{i \in I} |A_i|^T |O_i|^{T-1}$ integer variables) while **MILP-$n$ agents** has two 0-1 variables for every terminal as well as nonterminal history of each agent (*approximatively* $2 \sum_{i \in I} (|A_i||O_i|)^T$ integer variables).

### 5.5 Summary

The formulation of the solution of a DEC-POMDP and the application of the Duality Theorem for Linear Programs allow us to formulate the solution of a DEC-POMDP as the solution of a new kind of 0-1 MILP. For 2 agents, this MILP has “only” $O(k^T)$ variables and constraints and is thus “smaller” than **MILP** of the previous section. Still, all these MILPS are quite large and the next section investigates heuristic ways to speed up their resolution.

## 6. Heuristics for Speeding up the Mathematical Programs

This section focusses on ways to speed up the resolution of the various MILPs presented so far. Two ideas are exploited. First, we show how to prune the set of sequence-form policies by removing histories that will provably not be part of the optimal joint policy. These histories are called “locally extraneous”. Then, we give some lower and uppers bounds to the objective function of the MILPs, these bounds can sometimes be used in the “branch and bound” method often used by MILP solvers to finalize the values of the integer variables.

### 6.1 Locally Extraneous Histories

A locally extraneous history is a history that is not required to find an optimal joint policy when the initial state of the DEC-POMDP is $\alpha$ because it could be replaced by a *co-history* without affecting the value of the joint policy. A **co-history** of a history $h$ of an agent is defined to be a history of that agent that is identical to $h$ in all aspects *except* for its last action. If $A_i = \{b, c\}$, the only co-history of $c.u.b.v.b$ is the history $c.u.b.v.c$. The set of co-histories of a history $h$ shall be denoted by $C(h)$.

Formally, a history $h \in \mathcal{H}_i^t$ of length $t$ of agent $i$ is said to be **locally extraneous** if, for every probability distribution $\gamma$ over the set $\mathcal{H}_{-i}^t$ of $i$-reduced joint histories of length $t$, there exists a history $h' \in C(h)$ such that

$$\sum_{j' \in \mathcal{H}_{-i}^t} \gamma(j') \big\{ \mathcal{R}(\alpha, \langle h', j' \rangle) - \mathcal{R}(\alpha, \langle h, j' \rangle) \big\} \quad \geq \quad 0 \tag{130}$$

where $\gamma(j')$ denotes the probability of $j'$ in $\gamma$.

An alternative definition is as follows. A history $h \in \mathcal{H}_i^t$ of length $t$ of agent $i$ is said to be **locally extraneous** if there exists a probability distribution $\omega$ over the set of co-histories of $h$ such that for each $i$-reduced joint history $j'$ of length $t$, there holds

$$\sum_{h' \in C(h)} \omega(h') \mathcal{R}(\alpha, \langle h', j' \rangle) \geq \mathcal{R}(\alpha, \langle h, j' \rangle) \tag{131}$$

where $\omega(h')$ denotes the probability of the co-history $h'$ in $\omega$.

The following theorem justifies our incremental pruning of locally extraneous histories so that the search for optimal joint policies is faster because it is performed on a smaller set of possible support histories.

**Theorem 6.1.** *For every optimal $T$-period joint policy $p'$ such that for some agent $i \in I$ and for a terminal history $h$ of agent $i$ that is locally extraneous at $\alpha$, $p'_i(h) > 0$, there exists another $T$-period joint policy $p$ that is optimal at $\alpha$ and that is identical to $p'$ in all respects except that $p_i(h) = 0$.*

*Proof:* Let $p'$ be a $T$-period joint policy that is optimal at $\alpha$. Assume that for some agent $i \in I$ and for a terminal history $h$ of agent $i$ that is locally extraneous at $\alpha$, $p'_i(h) > 0$. By (130), there exists at least one co-history $h'$ of $h$ such that,

$$\sum_{j' \in \mathcal{H}^T_{-i}} p'_{-i}(j') \big\{ \mathcal{R}(\alpha, \langle h', j' \rangle) - \mathcal{R}(\alpha, \langle h, j' \rangle) \big\} \quad \geq \quad 0. \tag{132}$$

Let $q$ be a $T$-period policy of agent $i$ that is identical to $p'_i$ in all respects except that $q(h')$ $= p'_i(h) + p'_i(h')$ and $q(h) = 0$. We shall show that $q$ is also optimal at $\alpha$. There holds,

$$\begin{aligned} \mathcal{V}(\alpha, \langle q, p'_{-i} \rangle) - \mathcal{V}(\alpha, \langle p'_i, p_{-i} \rangle) &= \\ \sum_{j' \in \mathcal{H}^T_{-i}} p'_{-i}(j') \big\{ \mathcal{R}(\alpha, \langle h', j' \rangle) q(h') - \mathcal{R}(\alpha, \langle h', j' \rangle) p'_i(h') - \mathcal{R}(\alpha, \langle h, j' \rangle) p'_i(h) \big\} &= \\ \sum_{j' \in \mathcal{H}^T_{-i}} p'_{-i}(j') \big\{ \mathcal{R}(\alpha, \langle h', j' \rangle)(q(h') - p'_i(h')) - \mathcal{R}(\alpha, \langle h, j' \rangle) p'_i(h) \big\} &= \\ \sum_{j' \in \mathcal{H}^T_{-i}} p'_{-i}(j') \big\{ \mathcal{R}(\alpha, \langle h', j' \rangle) p'_i(h) - \mathcal{R}(\alpha, \langle h, j' \rangle) p'_i(h) \big\} & \end{aligned}$$

since $q(h') = p'_i(h) + p'_i(h')$. Therefore,

$$\begin{aligned} \mathcal{V}(\alpha, \langle q, p'_{-i} \rangle) - \mathcal{V}(\alpha, \langle p'_i, p_{-i} \rangle) &= \\ \sum_{j' \in \mathcal{H}^T_{-i}} p'_{-i}(j') \big\{ \mathcal{R}(\alpha, \langle h', j' \rangle) - \mathcal{R}(\alpha, \langle h, j' \rangle) \big\} &\geq 0 \quad \text{(due to (132))}. \end{aligned}$$

Hence, $p = \langle q,\, p'_{-i} \rangle$ is also an optimal $T$-period joint policy at $\alpha$. □

One could also wonder if the order with which extraneous histories are pruned is important or not. To answer this question, the following theorem shows that if many co-histories are extraneous, they can be pruned in any order as:

- either they all have the same value, so any one of them can be pruned ;
- or pruning one of them does not change the fact that the others are still extraneous.

**Theorem 6.2.** *If two co-histories $h_1$ and $h_2$ are both locally extraneous, either their values $\mathcal{R}(\alpha, \langle h_1, j' \rangle)$ and $\mathcal{R}(\alpha, \langle h_2, j' \rangle)$for all $j' \in \mathcal{H}^t_{-i}$ are equal or $h_1$ is also locally extraneous relatively to $C(h) \setminus \{h_2\}$.*

*Proof:* Let $C^+$ denotes the union $C(h_1) \cup C(h_2)$. We have immediately that $C(h_1) = C^+ \setminus \{h_1\}$ and $C(h_2) = C^+ \setminus \{h_2\}$. $h_1$ (resp. $h_2$) being locally extraneous means that there exists a probability distribution $\omega_1$ on $C(h_1)$ (resp. $\omega_2$ on $C(h_2)$) such that, for all $j'$ of $\mathcal{H}^t_{-i}$:

$$\sum_{h' \in C^+ \setminus \{h_1\}} \omega_1(h')\mathcal{R}(\alpha, \langle h', j' \rangle) \geq \mathcal{R}(\alpha, \langle h_1, j' \rangle) \tag{133}$$

$$\sum_{h' \in C^+ \setminus \{h_2\}} \omega_2(h')\mathcal{R}(\alpha, \langle h', j' \rangle) \geq \mathcal{R}(\alpha, \langle h_2, j' \rangle) \tag{134}$$

$$\tag{135}$$

Eq. (133) can be expanded in:

$$\omega_1(h_2)\mathcal{R}(\alpha, \langle h_2, j' \rangle) + \sum_{h' \in C^+ \setminus \{h_1, h_2\}} \omega_1(h')\mathcal{R}(\alpha, \langle h', j' \rangle) \geq \mathcal{R}(\alpha, \langle h_1, j' \rangle). \tag{136}$$

Using (134) in (136) gives

$$\omega_1(h_2) \sum_{h' \in C^+ \setminus \{h_2\}} \omega_2(h')\mathcal{R}(\alpha, \langle h', j' \rangle) + \sum_{h' \in C^+ \setminus \{h_1, h_2\}} \omega_1(h')\mathcal{R}(\alpha, \langle h', j' \rangle) \geq \mathcal{R}(\alpha, \langle h_1, j' \rangle) \tag{137}$$

leading to

$$\sum_{h' \in C^+ \setminus \{h_1, h_2\}} (\omega_1(h_2)\omega_2(h') + \omega_1(h'))\mathcal{R}(\alpha, \langle h', j' \rangle) \geq (1 - \omega_1(h_2)\omega_2(h_1))\mathcal{R}(\alpha, \langle h_1, j' \rangle) \tag{138}$$

So, two cases are possible:

- $\omega_1(h_2) = \omega_2(h_1) = 1$. In that case, as $\mathcal{R}(\alpha, \langle h_2, j' \rangle) \geq \mathcal{R}(\alpha, \langle h_1, j' \rangle)$ and $\mathcal{R}(\alpha, \langle h_1, j' \rangle) \geq \mathcal{R}(\alpha, \langle h_2, j' \rangle)$, we have that $\mathcal{R}(\alpha, \langle h_1, j' \rangle) = \mathcal{R}(\alpha, \langle h_2, j' \rangle)$ for all $j'$ of $\mathcal{H}^t_{-i}$.

- $\omega_1(h_2)\omega_2(h_1) < 1$. In that case we have:

$$\sum_{h' \in C^+ \setminus \{h_1, h_2\}} \frac{\omega_1(h_2)\omega_2(h') + \omega_1(h')}{1 - \omega_1(h_2)\omega_2(h_1)}\mathcal{R}(\alpha, \langle h', j' \rangle) \geq \mathcal{R}(\alpha, \langle h_1, j' \rangle) \tag{139}$$

  meaning that even without using $h_2$, $h_1$ is still locally extraneous because $\frac{\omega_1(h_2)\omega_2(h') + \omega_1(h')}{1 - \omega_1(h_2)\omega_2(h_1)}$ is a probability distribution over $C^+ \setminus \{h_1, h_2\}$

$$\sum_{h' \in C^+ \setminus \{h_1, h_2\}} \frac{\omega_1(h_2)\omega_2(h') + \omega_1(h')}{1 - \omega_1(h_2)\omega_2(h_1)} = \frac{\omega_1(h_2)(1 - \omega_2(h_1)) + (1 - \omega_1(h_2))}{1 - \omega_1(h_2)\omega_2(h_1)} \tag{140}$$

$$= \frac{1 - \omega_1(h_2)\omega_2(h_1)}{1 - \omega_1(h_2)\omega_2(h_1)} \tag{141}$$

$$= 1. \tag{142}$$

□

In order to prune locally extraneous histories, one must be able to identify these histories. There are indeed two complementary ways for doing this.

The first method relies on the definition of the value of a history (see Section 3.3), that is

$$\mathcal{R}(\alpha, \langle h, j' \rangle) \quad = \quad \Psi(\alpha, \langle h, j' \rangle)\overline{R}(\alpha, \langle h, j' \rangle). \tag{143}$$

Therefore, if

$$\Psi(\alpha, \langle h, j' \rangle) \quad = \quad 0, \quad \forall j' \in \mathcal{H}^t_{-i} \tag{144}$$

is true for a history $h$, then that means that every joint history of length $t$ occurring from $\alpha$ of which the given history is a part of has an *a priori* probability of 0. thus, $h$ is clearly extraneous. Besides, every co-history of $h$ will also be locally extraneous as they share the same probabilities.

A second test is needed because some locally extraneous histories do not verify (144). Once again, we turn to linear programing and in particular to the following linear program

Variables: $y(j)$, $\forall j \in \mathcal{H}^t_{-i}$

$$\text{Minimize} \quad \epsilon \tag{145}$$

subject to:

$$\sum_{j' \in \mathcal{H}^t_{-i}} y(j')\big\{\mathcal{R}(\alpha, \langle h', j' \rangle) - \mathcal{R}(\alpha, \langle h, j' \rangle)\big\} \quad \leq \quad \epsilon, \quad \forall h' \in C(h) \tag{146}$$

$$\sum_{j' \in \mathcal{H}^t_{-i}} y(j') \quad = \quad 1 \tag{147}$$

$$y(j') \quad \geq \quad 0, \quad \forall j' \in \mathcal{H}^t_{-i} \tag{148}$$

because of the following Lemma.

**Lemma 6.1.** *If, it exists a solution ($\epsilon^*$, $y^*$) to the linear program* (145)-(148) *where* $\epsilon^* \geq 0$*, then $h$ is locally extraneous.*

*Proof*: Let ($\epsilon^*$, $y^*$) be a solution to the LP (145)-(148). $y^*$ is a probability distribution over $\mathcal{H}^t_{-i}$ due to constraints (147)-(148). If $\epsilon^* \geq 0$, since we are minimizing $\epsilon$, due to constraints (146), we have that for every $\tilde{y} \in \Delta(\mathcal{H}^t_{-i})$, and for every co-history $h'$ of $h$

$$\sum_{j' \in \mathcal{H}^t_{-i}} \tilde{y}(j')\big\{\mathcal{R}(\alpha, \langle h', j' \rangle) - \mathcal{R}(\alpha, \langle h, j' \rangle)\big\} \quad \geq \quad \epsilon^*. \tag{149}$$

Therefore, by definition, $h$ is locally extraneous. □

The following procedure identifies all locally extraneous terminal histories of all the agents and proceed to their iterative pruning. This is mainly motivated by Theorems 6.1 and 6.2 for effectively removing extraneous histories. The procedure is similar to the procedure of iterated elimination of dominated strategies in a game (Osborne & Rubinstein, 1994). The concept is also quite similar to the process of policy elimination in the backward step of the dynamic programming for partially observable stochastic games (Hansen et al., 2004).

- **Step 1:** For each agent $i \in I$, set $\tilde{H}_i^T$ to $\mathcal{E}_i$. Let $\tilde{H}^T$ denote the set $\times_{i \in I} \tilde{H}_i^T$. For each joint history $j \in \tilde{H}^T$, compute and store the value $\mathcal{R}(\alpha, j)$ of $j$ and the joint observation sequence probability $\Psi(\alpha, j)$ of $j$.

- **Step 2:** For each agent $i \in I$, for each history $h \in \tilde{H}_i^T$, if for each $i$-reduced joint history $j' \in \tilde{H}_{-i}^T$, $\Psi(\alpha, \langle h, j' \rangle) = 0$, remove $h$ from $\tilde{H}_i^T$.

- **Step 3:** For each agent $i \in I$, for each history $h \in \tilde{H}_i^T$ do as follows: If $C(h) \cap \tilde{H}_i^T$ is non-empty, check whether $h$ is locally extraneous or not by setting up and solving LP (145)-(148). When setting the LP, replace $\mathcal{H}_{-i}^t$ by the set $\tilde{H}_{-i}^T$ and the set $C(h)$ by the set $C(h) \cap \tilde{H}_i^T$. If upon solving the LP, $h$ is found to be locally extraneous at $\alpha$, remove $h$ from $\tilde{H}_i^T$.

- **Step 4:** If in Step 3 a history (of any agent) is found to be locally extraneous, go to Step 3. Otherwise, terminate the procedure.

The procedure builds the set $\tilde{H}_i^T$ for each agent $i$. This set contains every terminal history of agent $i$ that is required for finding an optimal joint policy at $\alpha$, that is every terminal history that is not locally extraneous at $\alpha$. For each agent $i$, every history that is in $\mathcal{H}_i^T$ but not in $\tilde{H}_i^T$ is locally extraneous. The reason for reiterating Step 3 is that if a history $h$ of some agent $i$ is found to be locally extraneous and consequently removed from $\tilde{H}_i^T$, it is possible that a history of some other agent that was previously not locally extraneous now becomes so, due to the removal of $h$ from $\tilde{H}_i^T$. Hence, in order to verify if this is the case for any history or not, we reiterate Step 3.

Besides, Step 2 of the procedure below also prunes histories that are impossible given the model of the DEC-POMDP because their observation sequence can not be observed.

A last pruning step can be taken in order to remove non-terminal histories that can *only* lead to extraneous terminal histories. This last step is recursive, starting from histories of horizon $T-1$, we remove histories $h_i$ that have no non-extraneous terminal histories, that is, histories $h_i$ such that all $h.o.a$ are extraneous for $a \in A_i$ and $o \in O_i$.

**Complexity** The algorithm for pruning locally extraneous histories has an exponential complexity. Each joint history must be examined to compute its value and its occurence probability. Then, in the worst case, a Linear Program can be run for every local history in order to check it is extraneous or not. Experimentations are needed to see if the prunning is really interesting.

### 6.2 Cutting Planes

Previous heuristics were aimed at reducing the search space of the linear programs, which incidentally has a good impact on the time needed to solve these programs. Another option which directly aims at reducing the computation time is to use cutting planes (Cornuéjols, 2008). A cut (Dantzig, 1960) is a special constraint that identifies a portion of the set of feasible solutions in which the optimal solution provably does not lie. Cuts are used in conjunction with various "branch and bounds" mechanism to reduce the number of possibles combination of integer variables that are examined by a solver.

We will present two kinds of cuts.

Variables: $y(j)$, $\forall j \in \mathcal{H}$

$$\text{Maximize} \quad \sum_{j \in \mathcal{E}} \mathcal{R}(\alpha, j) y(j) \tag{153}$$

subject to,

$$\sum_{a \in A} y(a) = 1 \tag{154}$$

$$-y(j) + \sum_{a \in A} y(j.o.a) = 0, \quad \forall j \in \mathcal{N}, \forall o \in O \tag{155}$$

$$y(j) \geq 0, \quad \forall j \in \mathcal{H} \tag{156}$$

Table 6: **POMDP**. This linear program finds an optimal policy for a POMDP.

#### 6.2.1 Upper Bound for the Objective Function

The first cut we propose is the **upper bound POMDP cut**. The value of an optimal $T$-period joint policy at $\alpha$ for a given DEC-POMDP is bounded from above by the value $\mathcal{V}_P^*$ of an optimal $T$-period policy at $\alpha$ for the POMDP derived from the DEC-POMDP. This derived POMDP is the DEC-POMDP but assuming a centralized controller (i.e. with only one agent using joint-actions).

A sequence-form representation of the POMDP is quite straightforward. Calling $\mathcal{H}$ the set $\cup_{t=1}^{T} \mathcal{H}^t$ of joint histories of lengths less than or equal to $T$ and $\mathcal{N}$ the set $\mathcal{H} \backslash \mathcal{E}$ of non-terminal joint histories, a policy for POMDP with horizon $T$ in sequence-form is a function $q$ from $\mathcal{H}$ to $[0, 1]$ such that:

$$\sum_{a \in A} q(a) = 1 \tag{150}$$

$$-q(j) + \sum_{a \in A} q(j.o.a) = 0, \quad \forall j \in \mathcal{N}, \forall o \in O \tag{151}$$

The value $\mathcal{V}_P(\alpha, q)$ of a sequence-form policy $q$ is then given by:

$$\mathcal{V}_P(\alpha, q) = \sum_{j \in \mathcal{E}} \mathcal{R}(\alpha, j) q(j) \tag{152}$$

Thereby, the solution $y^*$ of the linear program of Table 6 is an optimal policy for the POMDP of horizon $T$ and the optimal value of the POMDP is $\sum_{j \in \mathcal{E}} \mathcal{R}(\alpha, j) y^*(j)$. So, the value $\mathcal{V}(\alpha, p^*)$ of the optimal joint policy $p^* = \langle p_1^*,\ p_2^*,\ \cdots,\ p_n^* \rangle$ of the DEC-POMDP is bounded by above by the value $\mathcal{V}_P(\alpha, q^*)$ of the associated POMDP.

**Complexity** The complexity of finding an upper bound is linked to the complexity of solving a POMDP which, as showed by Papadimitriou and Tsitsiklis (1987), can be PSPACE (*i.e.* require a memory that is polynomial in the size of the problem, leading to a possible exponential complexity in time). Once again, only experimentation can help us decide in which cases the upper bound cut is efficient.

#### 6.2.2 Lower Bound for the Objective Function

In the case of DEC-POMDPs with non-negative reward, it is trivial to show that the value of a $T$-period optimal policy is bounded from below by the value of the $T-1$ horizon optimal value. So, in the general case, we have to take into account the lowest reward possible to compute this lower bound and we can say that:

$$\sum_{j\in\mathcal{E}}\mathcal{R}(\alpha,j)z(j) \quad\geq\quad \mathcal{V}^{T-1}(\alpha)+\min_{a\in A}\min_{s\in S}R(s,a) \tag{157}$$

where $\mathcal{V}^{T-1}$ is the value of the optimal policy with horizon $T-1$. The reasoning leads to an iterated computation of DEC-POMDPs of longer and longer horizon, reminiscent of the MAA* algorithm (Szer et al., 2005). Experiments will tell if it is worthwhile to solve bigger and bigger DEC-POMDPs to take advantage of a lower bound or if it is better to directly tackle the $T$ horizon problem without using any lower bound.

**Complexity** To compute the lower bound, one is required to solve a DEC-POMDP whith an horizon that is one step shorter than the current horizon. The complexity is clearly at least exponential. In our experiments, the value of a DEC-POMDP has been used for the same DEC-POMDP with a bigger horizon. In such case, the computation time has been augmented by the best time to solve the smaller DEC-POMDP.

### 6.3 Summary

Pruning locally extraneous histories and using the bounds of the objective function can be of practical use for software solving the MILPs presented in this paper. Pruning histories means that the space of policies used by the MILP is reduced and, because the formulation of the MILP depends on combinatorial characteristics of the DEC-POMDP, these MILP must be altered as show in Appendix D.

**Validity** As far as cuts are concerned, they do not alter the solution found by the MILPs, so a solution to these MILPs is still an optimal solution to the DEC-POMDP. When extraneous histories are pruned, at least one valid policy is left as a solution because, in step 3 of the algorithm, an history is pruned only if it has other co-histories left. Besides, this reduced set of histories can still be used to build an optimal policy because of Theroem 6.1. As a consequence, the MILP build on this reduced set of histories admit a solution and this solution is one optimal joint policy.

In the next section, experimental results will allow us to understand in which cases the heuristics introduced can be useful.

## 7. Experiments

The mathematical programs and the heuristics designed in this paper are tested on four classical problems found in the literature. For these problems, involving *two* agents, we have mainly compared the computation time required to solve a DEC-POMDP using Mixed Integer Linear Programming methods to computation time reported for methods found in the literature. Then we have tested our programs on *three*-agent problems randomly designed.

| Problem | $\vert A_i\vert$ | $\vert O_i\vert$ | $\vert S\vert$ | $n$ |
|---|---|---|---|---|
| MABC | 2 | 2 | 4 | 2 |
| MA-Tiger | 3 | 2 | 2 | 2 |
| Fire Fighting | 3 | 2 | 27 | 2 |
| Grid Meeting | 5 | 2 | 16 | 2 |
| Random Pbs | 2 | 2 | 50 | 3 |

Table 7: "Complexity" of the various problems used as test beds.

**MILP** and **MILP-2** are solved using the "iLog Cplex 10" solver – a commercial set of Java packages – that relies on a combination of the "Simplex" and "Branch and Bounds" methods (Fletcher, 1987). The software is run on an Intel P4 at 3.4 GHz with 2Gb of RAM using default configuration parameters. For the mathematical programs, different combination of heuristics have been evaluated: pruning of locally extraneous histories, using a lower bound cut and using an upper bound cut, respectively denoted "LOC", "Low" and "Up" in the result tables to come.

The Non-Linear Program (**NLP**) of Section 3.4 has been evaluated by using various solvers from the NEOS website (`http://www-neos.mcs.anl.gov` ), even thought this method does not guarantee an optimal solution to the DEC-POMDP. Three solvers have been used: LANCELOT (abbreviated as LANC.), LOQO and SNOPT.

The result tables also report results found in the literature for the following algorithms: DP stands for Dynamic Programming from Hansen et al. (2004); DP-LPC is an improved version of Dynamic Programming where policies are compressed in order to fit more of them in memory and speed up their evaluation as proposed by Boularias and Chaib-draa (2008); PBDP is an extension of Dynamic Programming where pruning is guided by the knowledge of reachable belief-states as detailed in the work of Szer and Charpillet (2006); MAA* is a heuristically guided forward search proposed by Szer et al. (2005) and a generalized and improved version of this algorithm called GMAA* developed by Oliehoek et al. (2008).

The problems selected to evaluate the algorithms are detailed in the coming subsections. They have been widely used to evaluate DEC-POMDPs algorithms in the literature and their "complexity", in term of space size, is summarized in Table 7.

### 7.1 Multi-Access Broadcast Channel Problem

Several versions of the Multi-Access Broadcast Channel (MABC) problem can be found in the literature. We will use the description given by Hansen et al. (2004) that allows this problem to be formalized as a DEC-POMDP.

In the MABC, we are given two nodes (computers) which are required to send messages to each other over a common channel for a given duration of time. Time is imagined to be split into discrete periods. Each node has a buffer with a capacity of one message. A buffer that is empty in a period is refilled with a certain probability in the next period. In a period, only one node can send a message. If both nodes send a message in the same period, a collision of the messages occurs and neither message is transmitted. In case of a collision, each node is intimated about it through a collision signal. But the collision signaling mechanism is faulty. In case of a collision, with a certain probability, it does not send a signal to either one or both nodes.

We are interested in pre-allocating the channel amongst the two nodes for a given number of periods. The pre-allocation consists of giving the channel to one or both nodes in a period as a function of the node's information in that period. A node's information in a period consists only of the sequence of collision signals it has received till that period.

In modeling this problem as a DEC-POMDP, we obtain a 2-agent, 4-state, 2-actions-per-agent, 2-observations-per-agent DEC-POMDP whose components are as follows.

- Each node is an agent.
- The state of the problem is described by the states of the buffers of the two nodes. The state of a buffer is either Empty or Full. Hence, the problem has four states: (Empty, Empty), (Empty, Full), (Full, Empty) and (Full, Full).
- Each node has two possible actions, Use Channel and Don't Use Channel.
- In a period, a node may either receive a collision signal or it may not. So each node has two possible observations, Collision and No Collision.

The initial state of the problem $\alpha$ is (Full, Full). The state transition function $\mathbb{P}$, the joint observation function $\mathbb{G}$ and the reward function $R$ have been taken from Hansen et al. (2004). If both agents have full buffers in a period, and both use the channel in that period, the state of the problem is unchanged in the next period; both agents have full buffers in the next period. If an agent has a full buffer in a period and only he uses the channel in that period, then his buffer is refilled with a certain probability in the next period. For agent 1, this probability is 0.9 and for agent 2, this probability is 0.1. If both agents have empty buffers in a period, irrespective of the actions they take in that period, their buffers get refilled with probabilities 0.9 (for agent 1) and 0.1 (for agent 2).

The observation function $\mathbb{G}$ is as follows. If the state in a period is (Full, Full) and the joint action taken by the agents in the previous period is (Use Channel, Use Channel), the probability that both receive a collision signal is 0.81, the probability that only one of them receives a collision signal is 0.09 and the probability that neither of them receives a collision signal is 0.01. For any other state the problem may be in a period and for any other joint action the agents may have taken in the previous period, the agents do not receive a collision signal.

The reward function $R$ is quite simple. If the state in a period is (Full, Empty) and the joint action taken is (Use Channel, Don't Use Channel) or if the state in a period is (Empty, Full) and the joint action taken is (Don't Use Channel, Use Channel), the reward is 1; for any other combination of state and joint action, the reward is 0.

We have evaluated the various algorithms on this problem for three different horizons (3, 4 and 5) and the respective optimal policies have a value of 2.99, 3.89 and 4.79. Results are detailed in Table 8 where, for each horizon and algorithm, the value and the computation time for the best policy found are given.

The results show that the MILP compares favorably to more classical algorithms except for GMAA* that is always far better for horizon 4 and, for horizon 5, roughly within the same order of magnitude as **MILP** with the more pertinent heuristics. As expected, apart for the simplest setting (horizon of 3), **NLP** based resolution can not find the optimal policy of the DEC-POMDP, but the computation time is lower than the other methods. Among MILP methods, **MILP-2** is better than **MILP** even with the best heuristics for horizon 3 and 4. When the size of the problem increases, heuristics are the only way for MILPs to be able to cope with the size of the problem. The table also shows that, for the MABC problem, pruning extraneous histories using the LOC heuristic is always a good method and further investigation revealed that 62% of the heuristics proved to be locally extraneous. As far are cutting bounds are concerned, they don't seem to be very useful at first (for horizon 3 and 4) but are necessary for **MILP** to find a solution for horizon 5. For this problem, one must also have in mind that there is only one optimal policy for each horizon.

| Resolution method | | | Horizon 3 | | Horizon 4 | | Horizon 5 | |
|---|---|---|---|---|---|---|---|---|
| Program | Solver | Heuristics | Value | Time | Value | Time | Value | Time |
| **MILP** | Cplex | - | 2.99 | 0.86 | 3.89 | 900 | - | -m |
| **MILP** | Cplex | Low | 2.99 | 0.10 / 0.93 | 3.89 | 0.39 / 900 | - | 3.5 / -m |
| **MILP** | Cplex | Up | 2.99 | 0.28 / 1.03 | 3.89 | 0.56 / 907 | - | 4.73 / -m |
| **MILP** | Cplex | LOC | 2.99 | 0.34 / 0.84 | 3.89 | 1.05 / 80 | - | 2.27 / -t |
| **MILP** | Cplex | LOC, Low | 2.99 | 0.44 / 0.84 | 3.89 | 1.44 / 120 | - | 5.77 / -t |
| **MILP** | Cplex | LOC, Up | 2.99 | 0.62 / 0.93 | 3.89 | 1.61 / 10.2 | 4.79 | 7.00 / 25 |
| **MILP-2** | Cplex | - | 2.99 | 0.39 | 3.89 | 3.53 | - | -m |
| NLP | SNOPT | - | 2.90 | 0.01 | 3.17 | 0.01 | 4.70 | 0.21 |
| NLP | LANC. | - | 2.99 | 0.02 | 3.79 | 0.95 | 4.69 | 20 |
| NLP | LOQO | - | 2.90 | 0.01 | 3.79 | 0.05 | 4.69 | 0.18 |
| Algorithm | Family | | Value | Time | Value | Time | Value | Time |
| DP | Dyn. Prog. | | 2.99 | 5 | 3.89 | 17.59 | - | -m |
| DP-LPC | Dyn. Prog. | | **2.99** | **0.36** | 3.89 | 4.59 | - | -m |
| PBDP | Dyn. Prog. | | 2.99 | < 1s | 3.89 | 2 | 4.79 | $10^5$ |
| MAA* | Fw. Search | | 2.99 | < 1s | 3.89 | 5400 | - | -t |
| GMAA* | Fw. Search | | ? | ? | **3.89** | **0.03** | **4.79** | **5.68** |

Table 8: **MABC Problem**. Value and computation time (in seconds) for the solution of the problem as computed by several methods, best results are highlighted. When appropriate, time shows first the time used to run the heuristics then the global time, in the format `heuristic/total time`. "-t" means a timeout of 10,000s; "-m" indicates that the problem does not fit into memory and "?" indicates that the algorithm was not tested on that problem.

### 7.2 Multi-Agent Tiger Problem

As explained in section 2.2, the Multi-Agent Tiger problem (MA-Tiger) has been introduced in the paper from Nair et al. (2003). From the general description of the problem, we ob-tain a 2-agent, 2-state, 3-actions-per-agent, 2-observations-per agent DEC-POMDP whose elements are as follows.

| Joint Action | State | Joint Observation | Probability |
|---|---|---|---|
| (Listen, Listen) | Left | (Noise Left, Noise Left) | 0.7225 |
| (Listen, Listen) | Left | (Noise Left, Noise Right) | 0.1275 |
| (Listen, Listen) | Left | (Noise Right, Noise Left) | 0.1275 |
| (Listen, Listen) | Left | (Noise Right, Noise Right) | 0.0225 |
| (Listen, Listen) | Right | (Noise Left, Noise Left) | 0.0225 |
| (Listen, Listen) | Right | (Noise Left, Noise Right) | 0.1275 |
| (Listen, Listen) | Right | (Noise Right, Noise Left) | 0.1275 |
| (Listen, Listen) | Right | (Noise Right, Noise Right) | 0.7225 |
| (*, *) | * | (*, *) | 0.25 |

Table 9: Joint Observation Function $\mathbb{G}$ for the MA-Tiger Problem.

- Each person is an agent. So, we have a 2-agent DEC-POMDP.
- The state of the problem is described by the location of the tiger. Thus, $S$ consists of two states Left (tiger is behind the left door) and Right (tiger is behind the right door).
- Each agent's set of actions consists of three actions: Open Left (open the left door), Open Right (open the right door) and Listen (listen).
- Each agent's set of observations consists of two observations: Noise Left (noise coming from the left door) and Noise Right (noise coming from the right door).

The initial state is an equi-probability distribution over $S$. The state transition function $\mathbb{P}$, joint observation function $\mathbb{G}$ and the reward function $R$ are taken from the paper by Nair et al. (2003). $\mathbb{P}$ is quite simple. If one or both agents opens a door in a period, the state of the problem in the next period is set back to $\alpha$. If both agents listen in a period, the state of the process in unchanged in the next period. $\mathbb{G}$, given in Table (9), is also quite simple. Nair et al. (2003) describes two reward functions called "A" and "B" for this problem, here we report only results for reward function "A", given in Table 10, as the behavior of the algorithm are similar for both reward functions. The optimal value of this problem for horizons 3 and 4 are respectively 5.19 and 4.80.

For horizon 3, dynamic programming or forward search methods are generally better than mathematical programs. But this is the contrary for horizon 4 were the computation time of **MILP** with the "Low" heuristic is significatively better than any other, even GMAA*. Unlike MABC, the pruning of extraneous histories does not improve methods based on MILP, this is quite understandable as deeper investigations showed that there are *no* extraneous histories. Using lower cutting bounds proves to be very efficient and can be seen as a kind of heuristic search for the best policy ; not directly in the set of policies (like

| Joint Action | Left | Right |
|---|---|---|
| (Open Right, Open Right) | 20 | -50 |
| (Open Left, Open Left) | -50 | 20 |
| (Open Right, Open Left) | -100 | -100 |
| (Open Left, Open Right) | -100 | -100 |
| (Listen, Listen) | -2 | -2 |
| (Listen, Open Right) | 9 | -101 |
| (Open Right, Listen) | 9 | -101 |
| (Listen, Open Left) | -101 | 9 |
| (Open Left, Listen) | -101 | 9 |

Table 10: Reward Function "A" for the MA-Tiger Problem.

GMAA*) but in the set of combination of histories, which may explain the good behavior of **MILP**+Low.

It must also be noted that for this problem, approximate methods like **NLP** but also other algorithms not depicted here like the "Memory Bound Dynamic Programming" of Seuken and Zilberstein (2007) are able to find the optimal solution. And, once again, methods based on a NLP are quite fast and sometimes very accurate.

### 7.3 Fire Fighters Problem

The problem of the Fire Fighters (FF) has been introduced as a new benchmark by Oliehoek et al. (2008). It models a team of $n$ fire fighters that have to extinguish fires in a row of $n_h$ houses.

The state of each house is given by an integer parameter, called the fire level $f$, that takes discrete value between 0 (no fire) and $n_f$ (fire of maximum severity). At every time step, each agent can move to any one house. If two agents are at the same house, they extinguish any existing fire in that house. If an agent is alone, the fire level is lowered with a 0.6 probability if a neighbor house is also burning or with a 1 probability otherwise. A burning house with no fireman present will increase its fire level $f$ by one point with a 0.8 probability if a neighbor house is also burning or with a probability of 0.4 otherwise. An unattended non-burning house can catch fire with a probability of 0.8 if a neighbor house is burning. After an action, the agents receive a reward of $-f$ for each house that is still burning. Each agent can only observe if there are flames at its location with a probability that depends on the fire level: 0.2 if $f = 0$, 0.5 if $f = 1$ and 0.8 otherwise. At start, the agents are outside any of the houses and the fire level of the houses is sampled from a uniform distribution.

The model has the following characteristics:

- $n_a$ agents, each with $n_h$ actions and $n_f$ possible informations.
- There are $n_f^{n_h}.\binom{n_a+n_h-1}{n_a}$ states as there are $n_f^{n_h}$ possible states for the burning houses and $\binom{n_a+n_h-1}{n_a}$ different ways to distribute the $n_a$ fire fighters in the houses. For example, 2 agents with 3 houses and 3 levels of fire lead to $9 \times 6 = 54$ states. But, it is possible to use the information from the joint action to reduce the number of state needed in the transition function to simply $n_f^{n_h}$, meaning only 27 states for 2 agents with 3 houses and 3 levels of fire.

| **Resolution method** | | | Horizon 3 | | Horizon 4 | |
|---|---|---|---|---|---|---|
| Program | Solver | Heuristics | Value | Time | Value | Time |
| **MILP** | Cplex | - | 5.19 | 3.17 | - | -t |
| **MILP** | Cplex | Low | 5.19 | 0.46 / 4.9 | **4.80** | **3.5 / 72** |
| **MILP** | Cplex | Up | 5.19 | 0.42 / 3.5 | - | 0.75 / -t |
| **MILP** | Cplex | LOC | 5.19 | 1.41 / 6.4 | - | 16.0 / -t |
| **MILP** | Cplex | LOC, Low | 5.19 | 1.88 / 7.6 | 4.80 | 19.5 / 175 |
| **MILP** | Cplex | LOC, Up | 5.19 | 1.83 / 6.2 | - | 16.75 / -t |
| **MILP-2** | Cplex | - | 5.19 | 11.16 | - | -t |
| NLP | SNOPT | - | -45 | 0.03 | -9.80 | 4.62 |
| NLP | LANC. | - | 5.19 | 0.47 | 4.80 | 514 |
| NLP | LOQO | - | 5.19 | 0.01 | 4.78 | 91 |
| Algorithm | Family | | Value | Time | Value | Time |
| DP | Dyn. Prog. | | 5.19 | 2.29 | - | -m |
| DP-LPC | Dyn. Prog. | | 5.19 | 1.79 | 4.80 | 534 |
| PBDP | Dyn. Prog. | | ? | ? | ? | ? |
| MAA* | Fw. Search | | **5.19** | **0.02** | 4.80 | 5961 |
| GMAA* | Fw. Search | | 5.19 | 0.04 | 4.80 | 3208 |

Table 11: **MA-Tiger Problem**. Value and computation time (in seconds) for the solution of the problem as computed by several methods, best results are highlighted. When appropriate, time shows first the time used to run the heuristics then the global time, in the format `heuristic/total time`."-t" means a timeout of 10.000s; "-m" indicates that the problem does not fit into memory and "?" indicates that the algorithm was not tested on that problem.

| Resolution method | | | Horizon 3 | | Horizon 4 | |
|---|---|---|---|---|---|---|
| Program | Solver | Heuristics | Value | Time | Value | Time |
| **MILP** | Cplex | - | - | -t | - | -t |
| **MILP-2** | Cplex | - | -5.98 | 38 | - | -t |
| NLP | SNOPT | - | -5.98 | 0.05 | -7.08 | 4.61 |
| NLP | LANC. | - | -5.98 | 2.49 | -7.13 | 1637 |
| NLP | LOQO | - | -6.08 | 0.24 | -7.14 | 83 |
| Algorithm | Family | | Value | Time | Value | Time |
| MAA* | Fw. Search | | **(-5.73)** | **0.29** | (-6.57) | 5737 |
| GMAA* | Fw. Search | | (-5.73) | 0.41 | **(-6.57)** | **5510** |

Table 12: **Fire Fighting Problem**. Value and computation time (in seconds) for the solution of the problem as computed by several methods, best results are highlighted. "-t" means a timeout of 10.000s. For MAA* and GMAA*, value in parenthesis are taken from the work of Oliehoek et al. (2008) and *should* be optimal but are different from *our* optimal values.

- Transition, observation and reward functions are easily derived from the above description.

For this problem, dynamic programming based methods are not tested as the problem formulation is quite new. For horizon 3, the value of the optimal policy given by Oliehoek et al. (2008) ($-5.73$) differs from the value found by the MILP algorithms ($-5.98$) whereas both methods are supposed to be exact. This might come from slight differences in our respective formulation of the problems. For horizon 4, Oliehoek et al. (2008) report an optimal value of ($-6.57$).

For this problem, MILP methods are clearly outperformed by MAA* and GMAA*. Only NLP methods, which give an optimal solution for horizon 3, are better in term of computation time. It might be that NLP are also able to find optimal policies for horizon 4 but as our setting differs from the work of Oliehoek et al. (2008), we are not able to check if the policy found is really the optimal. The main reason for the superiority of forward search method lies in the fact that this problem admits many many optimal policies with the same value. In fact, for horizon 4, MILP-based methods find an optimal policy quite quickly (around 82s for **MILP-2**) but then, using branch-and-bound, must evaluate all the other potential policies before knowing that it indeed found an optimal policy. Forward search methods stop nearly as soon as they hit one optimal solution.

Heuristics are not reported as, not only do they not improve the performance of MILP but they take away some computation time and thus the results are worse.

### 7.4 Meeting on a Grid

The problem called "Meeting on a grid" deals with two agents that want to meet and stay together in a grid world. It has been introduced in the work of Bernstein, Hansen, and Zilberstein (2005).

In this problem, we have two robots navigating on a two-by-two grid world with no obstacles. Each robot can only sense whether there are walls to its left or right, and the goal is for the robots to spend as much time as possible on the same square. The actions are to move up, down, left or right, or to stay on the same square. When a robot attempts to move to an open square, it only goes in the intended direction with probability 0.6, otherwise it randomly either goes in another direction or stays in the same square. Any move into a wall results in staying in the same square. The robots do not interfere with each other and cannot sense each other. The reward is 1 when the agents share a square, and 0 otherwise. The initial state distribution is deterministic, placing both robots in the upper left corner of the grid.

The problem is modelled as a DEC-POMDP where:

- There are 2 agents, each one with 5 actions and  observations (wall on left, wall on right).
- There are 16 states, since each robot can be in any of 4 squares at any time.
- Transition, observation and reward functions are easily derived from the above description.

For this problem, dynamic programming based methods are not tested as the problem formulation is quite new. This problem is intrinsically more complex that FF and as such is only solved for horizon 2 and 3. Again, optimal value found by our method differ from the value reported by Oliehoek et al. (2008). Whereas we found that the optimal values are 1.12 and 1.87 for horizon 2 and 3, they report optimal values of 0.91 and 1.55.

Results for this problem have roughly the same pattern that the results for the FF problem. MAA* and GMAA* are quicker than MILP, but this time **MILP** is able to find an optimal solution for horizon 3. NLP methods give quite good results but they are slower than GMAA*. As for the FF, there are numerous optimal policies and MILP methods are not able to detect that the policy found quickly is indeed optimal.

Again, heuristics are not reported as, not only do they not improve the performance of MILP but they take away some computation time and thus the results are worse.

### 7.5 Random 3-Agent Problems

To test our approach on problems with 3 agents, we have used randomly generated DEC-POMDPs where the state transition function, the joint observation function and the reward functions are randomly generated. The DEC-POMDPs have 2 actions and 2 observations per agent and 50 states. Rewards are randomly generated integers in the range 1 to 5. The complexity of this family of problem is quite similar to the complexity of the MABC problem (see Section 7.1).

| Resolution method | | | Horizon 2 | | Horizon 3 | |
|---|---|---|---|---|---|---|
| Program | Solver | Heuristics | Value | Time | Value | Time |
| **MILP** | Cplex | - | 1.12 | 0.65 | 1.87 | 1624 |
| **MILP-2** | Cplex | - | 1.12 | 0.61 | - | -t |
| NLP | SNOPT | - | 0.91 | 0.01 | 1.26 | 1.05 |
| NLP | LANC. | - | 1.12 | 0.06 | 1.87 | 257 |
| NLP | LOQO | - | 1.12 | 0.07 | 0.48 | 81 |
| Algorithm | Family | | Value | Time | Value | Time |
| MAA* | Fw. Search | | **(0.91)** | **0s** | (1.55) | 10.8 |
| GMAA* | Fw. Search | | **(0.91)** | **0s** | **(1.55)** | **5.81** |

Table 13: **Meeting on a Grid Problem**. Value and computation time (in seconds) for the solution of the problem as computed by several methods, best results are highlighted. "-t" means a timeout of 10.000s. For MAA* and GMAA*, value in parenthesis are taken from the work of Oliehoek et al. (2008) and *should* be optimal but are different from *our* optimal values...

| Program | Least Time (secs) | Most Time (secs) | Average | Std. Deviation |
|---|---|---|---|---|
| **MILP** | 2.45 | 455 | 120.6 | 183.48 |
| **MILP-2** | 6.85 | 356 | 86.88 | 111.56 |

Table 14: Times taken by **MILP** and **MILP-2** on the 2-agent Random Problem for horizon 4.

In order to assess the "real" complexity of this Random problem, we have first tested a two-agent version of the problem for a horizon of 4. Results averaged over 10 runs of the programs are given in Table 14. When compared to the MABC problem which seemed of comparable complexity, the Random problem proves easier to solve (120s vs 900s). For this problem, the number of 0-1 variable is relatively small, as such it does not weight too much on the resolution time of **MILP-2** which is thus faster.

Results for a three-agent problem with horizon 3 are given in Table 15, once again averaged over 10 runs. Even though the size of the search space is "smaller" in that case (for 3 agents and a horizon of 3, there are $9 \times 10^{21}$ policies whereas the problem with 2 agents and horizon 4, there are $1.5 \times 10^{51}$ possible policies), the 3 agent problems seems more difficult to solve, demonstrating that one of the big issue is policy coordination. Here, heuristics bring a significative improvement on the resolution time of **MILP**. As predicted, **MILP-n** is not very efficient and is only given for completeness.

| Program | Least Time (secs) | Most Time (secs) | Average | Std. Deviation |
|---|---|---|---|---|
| **MILP** | 21 | 173 | 70.6 | 64.02 |
| **MILP**-Low | 26 | 90 | 53.2 | 24.2 |
| **MILP-n** | 754 | 2013 | 1173 | 715 |

Table 15: Times taken by **MILP** and **MILP-n** on the 3-agent Random problem for horizon 3.

## 8. Discussion

We have organized the discussion in two parts. In the first part, we analyze our results and offer explanations on the behavior of our algorithms and the usefulness of heuristics. Then, in a second part, we explicitely address some important questions.

### 8.1 Analysis of the Results

From the results, it appears that MILP methods are a better alternative to Dynamic Programming methods for solving DEC-POMDPs but are globally and generally clearly outperformed by forward search methods. The structure and thus the characteristics of the problem have a big influence on the efficiency of the MILP methods. Whereas it seems that the behavior of GMAA* in terms of computation time is quite correlated with the complexity of the problem (size of the action and observation spaces), MILP methods seem sometimes less correlated to this complexity. It is the case for the MABC problem (many extraneous histories can be pruned) and the MA-Tiger problem (special structure) where they outperform GMAA*. On the contrary, when many optimal policies exists, forward search methods like GMAA* are clearly a better choice. Finally, Non-Linear Programs, even though they can not guarantee an optimal solution, are generally a good alternative as they are sometimes able to find a very good solution and their computation time is often better than GMAA*. This might prove useful for approximate heuristic-driven forward searches.

The computational record of the two 2-agent programs shows that **MILP-2 agents** is slower than **MILP** when the horizon grows. There are two reasons to which the sluggishness of **MILP-2 agents** may be attributed. The time taken by the branch and bound (BB) method to solve a 0-1 MILP is inversely proportional to the number of 0-1 variables in the MILP. **MILP-2 agents** has many more 0-1 variables than **MILP** event hough the total number of variables in it is exponentially less than in **MILP**. This is the first reason. Secondly, **MILP-2 agents** is a more *complicated* program than **MILP**; it has many more constraints than **MILP**. **MILP** is a simple program, concerned only with finding a subset of a given set. In addition to finding weights of histories, **MILP** also finds weights of terminal joint histories. This is the only extra or superfluous quantity it is forced to find. On the other hand, **MILP-2 agents** takes a much more circuitous route, finding many more superfluous quantities than **MILP**. In addition to weights of histories, **MILP-2 agents** also finds supports of policies, regrets of histories and values of information sets. Thus, the relaxation of **MILP-2 agents** takes longer to solve than the relaxation of **MILP**. This is the second reason for the slowness with which the BB method solves **MILP-2 agents**.

| Problem | Heuristic | Horizon 2 | | Horizon 3 | | Horizon 4 | | Horizon 5 | |
|---|---|---|---|---|---|---|---|---|---|
| | | Time | #pruned | Time | #pruned | Time | #pruned | Time | #pruned |
| MABC | LOC | | | 0.34 | 14/32 | 1.047 | 74/128 | 2.27 | 350/512 |
| | Low | | | 0.10 | | 0.39 | | 3.5 | |
| | Up | | | 0.28 | | 3.89 | | 4.73 | |
| MA-Tiger | LOC | 0.41 | 0/18 | 1.41 | 0/108 | 16.0 | 0/648 | | |
| | Low | | | 0.46 | | 3.5 | | | |
| | Up | | | 0.42 | | 0.75 | | | |
| Meeting | LOC | 1.36 | 15/50* | 74.721 | 191/500* | | | | |

Table 16: **Computation time of heuristics**. For the LOC heuristics, we give the computation time in seconds and the number of locally extraneous histories pruned over the total number of histories (for an agent). A '*' denotes cases where *one* additional history is prunned for the second agent. For the Low and Up heuristic, only computation time is given.

For bigger problems, namely Fire-Fighters and Meeting on a Grid, when the horizon stays small, **MILP-2 agents** can compete with **MILP** because of its slightly lower size. Its complexity grows like $O((|A_i||O_i|)^T)$ whereas it grows like $O((|A_i||O_i|)^{2T})$ for **MILP**. But that small difference does not hold long as the number of integer variables quickly lessens the efficiency of **MILP-2 agents**.

As far as heuristic are concerned, they proved to be invaluable for some problems (MABC and MA-Tiger) and useless for others. In the case of MABC, heuristics are very helpful to prune a large number of extraneous heuristics but ultimately, it is the combination with the upper bound cut that it the more efficient when the horizon grows. In the case of MA-Tiger, although no extraneous histories are found, using the lower bound cut heuristic with **MILP** leads to the quickest algorithm for solving the problem with a horizon of 4. For other problems, heuristics are more of a burden as they are too greedy in computation time to speed up the resolution. For example, for the "Grid Meeting" problem, the time taken to prune extraneous histories is bigger than the time saved for solving the problem.

As a result, the added value of using heuristics depends on the nature of the problem (as depicted in Table 16) but, right now, we are not able to predict their usefulness without trying them.

We also emphasize that the results given here lie at the limit of what is possible to solve in an exact manner given the memory of the computer used for the resolution, especially in terms of the horizon. Furthermore, as the number of agent increases, the length of the horizon must be decreased for the problems to still be solvable.

### 8.2 Questions

The mathematical programing approach presented in this paper raises different questions. We have explicitly addressed some of the questions that appears important to us.

**Q1: Why is the sequence-form approach not entirely doomed by its exponential complexity?**

As the number of sequence-form *joint* policies grows doubly exponentially with the horizon and the number of agents, the sequence-form approach seems doomed, even compared to dynamic programming which is doubly exponential in the worst cases only. But, indeed, some arguments must be taken into consideration.

"Only" an exponential number of *individual* histories need to be evaluated. The "joint" part of the sequence-form is left to the MILP solver. And every computation done on a particular history, like computing its value or checking if it is extraneous, has a greater "reusability" than computations done on entire policies. An history is shared by many more joint policies than an individual policy. In some way, sequence-form allows us to work on reusable part of policies without having to work directly in the world of distributions on the set of joint-policies.

Then, the MILPs derived from the sequence-form DEC-POMDPs need a memory size which grows "only" exponentially with the horizon and the number of agents. Obviously, such a complexity is quickly overwhelming but it is also the case of every other exact method so far. As shown by the experiments, the MILP approach derived from the sequence-form compares quite well with dynamic programming, even if outperformed by forward methods like GMAA*.

**Q2: Why does MILP sometimes take so little time to find an optimal joint policy when compared to existing algorithms?**

Despite the complexity of our MILP approach, three factors contribute to the relative efficiency of **MILP**.

1. First, the efficiency of linear programming tools themselves. In solving **MILP**, the BB method solves a sequence of linear programs using the simplex algorithm. Each of these LPs is a relaxation of **MILP**. In theory, the simplex algorithm requires in the worst case an exponential number of steps (in the size of the LP) in solving a LP[2], but it is well known that, in practice, it usually solves a LP in a polynomial number of steps (in the size of the LP). Since the size of a relaxation of **MILP** is exponential in the horizon, this means that, roughly speaking, the time taken to solve a relaxation of **MILP** is "only" exponential in the horizon whereas it can be doubly exponential for other methods.

2. The second factor is the sparsity of the matrix of coefficients of the constraints of **MILP**. The sparsity of the matrix formed by the coefficients of the constraints of

2. This statement must be qualified: this worst case time requirement has not been demonstrated for all *variants* of the simplex algorithm. It has been demonstrated only for the basic version of the simplex algorithm.

an LP determines in practice the rate with which a pivoting algorithm such as the simplex solves the LP (this also applies to Lemke's algorithm in the context of an LCP). The sparser this matrix, the lesser the time required to perform elementary pivoting (row) operations involved in the simplex algorithm and the lesser the space required to model the LP.

3. The third factor is the fact that we supplement **MILP** with cuts; the computational experience clearly shows how this speeds up the computations. While the first two factors were related to solving a relaxation of **MILP** (i.e., an LP), this third factor has an impact on the BB method itself. The upper bound cut identifies an additional terminating condition for the BB method, thus enabling it to terminate earlier than in the absence of this condition. The lower bound cut attempts to shorten the list of active subproblems (LPs) which the BB method solves sequentially. Due to this cut, the BB method has potentially a lesser number of LPs to solve. Note that in inserting the lower bound cut, we are emulating the forward search properties of the A* algorithm.

**Q3: How do we know that the MILP-solver (iLog's "Cplex" in our experiments) is not the only reason for the speedup?**

Clearly, our approach would be slower, even sometime slower than a classical dynamic programming approach if we had used another program for solving our MILPs as we experimented also our MILPs with solvers from the NEOS website that were indeed very very slow. It is true that Cplex, the solver we have used in our experiments, is quite optimized. Nevertheless, it is exactly one of the points we wanted to experiment with in this paper: one of the advantages of formulating a DEC-POMDP as a MILP is the possibility to use the fact that, as mixed integer linear programs are very important for the industrial world, optimized solvers *do* exist.
Then, we *had* to formulate a DEC-POMDP as a MILP and this is mostly what this paper is about.

**Q4: What is the main contribution of this paper?**

As stated earlier in the paper, current algorithms for DEC-POMDPs were largely inspired by POMDPs algorithms. Our main contribution was to pursue an entirely different approach, *i.e.*, mixed integer linear programming. As such, we have learned a lot about DEC-POMDPs and about the *pro & con* of this mathematical programming approach. This has lead to the formulation of new algorithms.

In designing these algorithms, we have, first of all, drawn attention to a new representation of a policy, namely the sequence form of a policy, introduced by Koller, Megiddo and von Stengel. The **sequence form of a policy** is *not* a compact representation of the policy of an agent, but it does afford a compact representation of the *set* of policies of the agent.

The algorithms we have proposed for finite horizon DEC-POMDPs are **mathematical programming algorithms**. To be precise, they are 0-1 MILPs. In the MDP domain, mathematical programming has been long used for solving the infinite horizon case. For instance, an infinite horizon MDP can be solved by a linear program (d'Epenoux, 1963). More recently, mathematical programming has been directed at infinite horizon POMDPs and DEC-POMDPs. Thus, an infinite horizon DEC-MDP (with state transition independence) can be solved by a 0-1 MILP (Petrik & Zilberstein, 2007) and an infinite horizon POMDP or DEC-POMDP can be solved (for local optima) by a nonlinear program (Amato, Bernstein, & Zilberstein, 2007b, 2007a). The finite horizon case – much different in character than the infinite horizon case – has been dealt with using dynamic programming. As stated earlier, whereas dynamic programming has been quite successful for finite horizon MDPs and POMDPs, it has been less so for finite horizon DEC-POMDPs.

In contrast, in game theory, mathematical programming has been successfully directed at games of finite horizon. Lemke's algorithm (1965) for two-player normal form games, the Govindan-Wilson algorithm (2001) for $n$-player normal form games and the Koller, Megiddo and von Stengel approach (which internally uses Lemke's algorithm) for two-player extensive form games are all for finite-horizon games.

What remained then was to a find way to appropriate mathematical programming for solving the finite horizon case of the POMDP/DEC-POMDP domain. Our work has done precisely this (incidently, we now have an algorithm for solving some kind of $n$-player normal form games). Throughout the paper, we have shown how mathematical programming (in particular, 0-1 integer programming) can be applied for solving finite horizon DEC-POMDPs (it is easy to see that the approach we have presented yields a linear program for solving a finite horizon POMDP). Additionally, the computational experience of our approach indicates that for finite horizon DEC-POMDPs, mathematical programming may be better (faster) than dynamic programming. We have also shown how the well-entrenched dynamic programming heuristic of the pruning of redundant or extraneous objects (in our case, histories) can be integrated into this mathematical programming approach.
Hence, the main contribution of this paper is that it presents, for the first time, an alternative approach for solving finite horizon POMDPs/DEC-POMDPs based on MILPs.

**Q5: Is the mathematical programming approach presented in this paper something of a dead end?**

This question is bit controversial and a very short answer to this question could be a "small yes". But this is true for every approach that looks for *exact* optimal solutions to DEC-POMDPs, whether it is grounded on dynamic programming or forward search or mathematical programming. Because of the complexity of the problem, an exact solution will always be untractable but our algorithms can still be improved.

A longer answer is more mitigated, especially in the light of the recent advances made for dynamic programming and forward search algorithms. One crucial point in sequence-form DEC-POMDPs is the pruning of extraneous histories. A recent work from Oliehoek, Whiteson, and Spaan (2009) has shown how to clusters histories that are equivalent in a way that could also reduce the nomber of constraints in MILPs. The approach of Amato, Dibangoye, and Zilberstein (2009) that improves and speed up the dynamic programming operator could help in finding extraneous histories. So, at the very least, some work is still required before stating that every aspect of sequence-form DEC-POMDPs have been studied.

We now turn to an even longer answer. Consider the long horizon case. Given that exact algorithms (including the ones presented in this paper) can only tackle horizons less than 6, by 'long horizon', we mean anything upwards of 6 time periods. For the long horizon case, we are required to conceive a possibly sub-optimal joint policy for the given horizon and determine an upper bound on the loss of value incurred by using the joint policy instead of using an optimal joint policy.

The current trend for the long horizon case is a *memory-bounded* approach. The memory bounded dynamic programming (MBDP) algorithm (Seuken & Zilberstein, 2007) is the main exponent of this approach. This algorithm is based on the backward induction DP algorithm (Hansen et al., 2004). The algorithm attempts to run in a limited amount of space. In order to do so, unlike the DP algorithm, it prunes even non-extraneous (i.e., non-dominated) policy trees at each iteration. Thus, at each iteration, the algorithm retains a pre-determined number of trees. This algorithm and its variants have been used to find a joint policy for the MABC, the MA-tiger and the Box pushing problems for very long horizons (of the order of thousands of time periods).

MBDP does not provide an upper bound on the loss of value. The bounded DP (BDP) algorithm presented in the paper by Amato, Carlin, and Zilberstein (2007c) does give an upper bound. However, on more interesting DEC-POMDP problems (such as MA-tiger), MBDP finds a much better joint policy than BDP.

A meaningful way to introduce the notion of memory boundedness into our approach is to fix an *a priori* upper bound on the size of the concerned mathematical program. This presents all sorts of difficulties but the main difficulty seems to be the need to represent a policy for a long horizon in limited space. The MBDP algorithm solves this problem by using what may be termed as a recursive representation. The recursive representation causes the MBDP algorithm to take a long time to evaluate a joint policy, but it does allow the algorithm to represent a long horizon joint policy in limited space. In the context of our mathematical programming approach, we would have to change the policy constraints in some way so that a long horizon policy is represented by a system consisting of a limited number of linear equations and linear inequalities. Besides the policy constraints, other constraints of the presented programs would also have to be accordingly transfigured. It is not evident (to us) if such a transfiguration of the constraints is possible.

On the other hand, the infinite horizon case seems to be a promising candidate to adapt our approach to. Mathematical programming has already been applied, with some success, to solving infinite horizon DEC-POMDPs (Amato et al., 2007a). The computational experience of this mathematical programming approach shows that it is better (finds higher quality solutions in lesser time) than a dynamic programming approach (Bernstein et al., 2005; Szer & Charpillet, 2006).

Nevertheless, this approach has two inter-related shortcomings. First, the approach finds a joint controller (i.e., an infinite horizon joint policy) of a *fixed size* and not of the optimal size. Second, much graver than the first, for the fixed size, it finds a locally optimal joint controller. The approach does not guarantee finding an optimal joint controller. This is because the program presented in the work of Amato et al. (2007a) is a (non-convex) nonlinear program (NLP). The NLP finds a fixed size joint controller in the canonical form (i.e., in the form of a finite state machine). We believe that both these shortcomings can be removed by conceiving a mathematical program (specifically, a 0-1 mixed integer linear program) that finds a joint controller in the sequence-form. As stated earlier, the main challenge in this regard is therefore an identification of the sequence-form of an infinite horizon policy. In fact, it may be that if such sequence-form characterization of an infinite horizon policy is obtained, it could be used in conceiving a program for the long horizon (undiscounted reward) case as well.

**Q6: How does this help achieve designing artificial autonomous agents ?**

At first sight, our work does not have any direct and immediate applied benefits for the purpose of building artificial intelligent agents or understanding how intelligence "works". Even in the limited field of multi-agent planning, our contributions are more on a theoretical level than on a practical one.

Real artificial multi-agent systems can indeed be modeled as DEC-POMDPs, even if they make use of communication, of common knowledge, of common social law. Then, such real systems would likely be made of a large number of states, actions or observations and require solutions over a large horizon. Our mathematical programming approach is practically useless in that setting as limited to DEC-POMDPs of very small size. Other models that are simpler – but far from trivial – to solve because they explicitly take into account some characteristics of the real systems do exist. Some works take advantage of communications (Xuan, Lesser, & Zilberstein, 2000; Ghavamzadeh & Mahadevan, 2004), some of the existing independencies in the system (Wu & Durfee, 2006; Becker, Zilberstein, Lesser, & Goldman, 2004), some do focus on interaction between agents (Thomas, Bourjot, & Chevrier, 2004), some, as said while answering the previous questions, rely on approximate solutions, *etc...* It is our intention to facilitate the re-use and the adaptation to these other models of the concepts used in our work and of the knowledge about the structure of an optimal solution of a DEC-POMDP. To that end, we decided not only to describe the MILP programs but also, and most importantly, *how* we derived these programs by making use of some properties of optimal DEC-POMDP solutions.

Truly autonomous agents will also require to adapt to new and unforeseen situations. Our work being dedicated to planning, it seems easy to argue that it does not contribute very much to that end either. On the other hand, learning in DEC-POMDPs has never really been addressed except for some fringe work in particular settings (Scherrer & Charpillet, 2002; Ghavamzadeh & Mahadevan, 2004; Buffet, Dutech, & Charpillet, 2007). In fact, even for "simple" POMDPs, learning is a very difficult task (Singh, Jaakkola, & Jordan, 1994). Currently, the more promising research deals with learning the "Predictive State Representation" (PSR) of a POMDP (Singh, Littman, Jong, Pardoe, & Stone, 2003; James & Singh, 2004; McCracken & Bowling, 2005). Making due allowance to the fundamental differences between the functional role of PSR and histories, we notice that PSR and histories are quite similar in structure. While it is too early to say, it might be that trying to learn the useful histories of a DEC-POMDP could take some inspiration from the way the right PSRs are learned for POMDPs.

## 9. Conclusion

We designed and investigated new exact algorithms for solving Decentralized Partially Observable Markov Decision Processes with finite horizon (DEC-POMDPs). The main contribution of our paper is the use of sequence-form policies, based on a sets of histories, in order to reformulate a DEC-POMDP as a non-linear programming problem (**NLP**). We have then presented two different approaches to linearize the **NLP** in order to find global and optimal solutions to DEC-POMDPs. The first approach is based on the combinatorial properties of the optimal policies of DEC-POMDPs and the second one relies on concepts borrowed from the field of game theory. Both lead to formulating DEC-POMDPs as *0-1 Mixed Integer Linear Programming problems* (MILPs). Several heuristics for speeding up the resolution of these MILPs make another important contribution of our work.

Experimental validation of the mathematical programming problems designed in this work was conducted on classical DEC-POMDP problems found in the literature. These experiments show that, as expected, our MILP methods outperform classical Dynamic Programming algorithms. But, in general, they are less efficient and more costly than forward search methods like GMAA*, especially in the case where the DEC-POMDP admits many optimal policies. Nevertheless, according to the nature of the problem, MILP methods can sometimes greatly outperform GMAA* (as in the MA-Tiger problem).

While it is clear that exact resolution of DEC-POMDPs can not scale up with the size of the problems or the length of the horizon, designing exact methods is useful in order to develop or improve approximate methods. We see at least three research directions where our work can contribute. One direction could be to take advantage of the large literature on algorithms for finding approximate solutions to MILPs and to adapt them to the MILPs formulated for DEC-POMDPs. Another direction would be to use the knowledge gained from our work to derive improved heuristics for guiding existing approximate existing methods for DEC-POMDPs. For example, the work of Seuken and Zilberstein (2007), in order to limit the memory resources used by the resolution algorithm, prune the space of policies to only consider some of them; our work could help using a better estimation of the policies that are important to be kept in the search space. Then, the one direction we are currently investigating is to adapt our approach to DEC-POMDPs of infinite length by looking for yet another representation that would allow such problems to be seen as MILPs.

More importantly, our work participates to a better understanding of DEC-POMDPs. We analyzed and understood key characteristics of the nature of optimal policies in order to design the MILPs presented in this paper. This knowledge can be useful for other work dealing with DEC-POMDPs and even POMDPs. The experimentations have also given some interesting insights on the nature of the various problems tested, in term of existence of extraneous histories or on the number of optimal policies. These insights might be a first step toward a taxonomy of DEC-POMDPs.

## Appendix A. Non-Convex Non-Linear Program

Using the simplest example, this section aims at showing that the Non-Linear Program (**NLP**) expressed in Table 2 can be non-convex.

Let us consider an example with two agents, each one with 2 possible actions ($a$ and $b$) that want to solve a horizon-1 decision problem. The set of possible joint-histories is then: $\langle a, a\rangle$, $\langle a, b\rangle$, $\langle b, a\rangle$ and $\langle b, b\rangle$. Then the NLP to solve is:

Variables: $x_1(a)$, $x_1(b)$, $x_2(a)$, $x_2(a)$

$$\begin{aligned} \text{Maximize} \quad & \mathcal{R}(\alpha, \langle a, a\rangle)x_1(a)x_2(a) + \mathcal{R}(\alpha, \langle a, b\rangle)x_1(a)x_2(b) \\ & + \mathcal{R}(\alpha, \langle b, a\rangle)x_1(b)x_2(a) + \mathcal{R}(\alpha, \langle b, b\rangle)x_1(b)x_2(b) \end{aligned} \tag{158}$$

subject to

$$\begin{aligned} x_1(a) + x_1(b) &= 1 \\ x_2(a) + x_2(b) &= 1 \\ x_1(a) \geq 0, & \quad x_1(b) \geq 0 \\ x_2(a) \geq 0, & \quad x_2(b) \geq 0 \end{aligned}$$

A matrix formulation of the objective function of eq. (158) would be $x^T.C.x$ with $C$ and $x$ of the following kind:

$$C = \begin{bmatrix} 0 & 0 & c & d \\ 0 & 0 & e & f \\ c & e & 0 & 0 \\ d & f & 0 & 0 \end{bmatrix} \qquad x = \begin{bmatrix} x_1(a) \\ x_1(b) \\ x_2(a) \\ x_2(b) \end{bmatrix}. \tag{159}$$

If $\lambda$ is the eigen value of vector $v = [v_1\ v_2\ v_3\ v_4]^T$ then it is straightforward to show that $-\lambda$ is also an eigen value: $[-v_1\ -v_2\ v_3\ v_4]^T = -\lambda C.[v_1\ v_2\ -v_3\ -v_4]^T$. As a result, the matrix $C$, hessian of the objective function, is not positive-definite and thus the objective function is not convex.

## Appendix B. Linear Program Duality

Every linear program (LP) has a converse linear program called its dual. The first LP is called the primal to distinguish it from its dual. If the primal maximizes a quantity, the dual minimizes the quantity. If there are $n$ variables and $m$ constraints in the primal, there are $m$ variables and $n$ constraints in the dual. Consider the following (primal) LP.

Variables: $x(i)$, $\forall i \in \{1, 2, \cdots, n\}$

$$\text{Maximize} \quad \sum_{i=1}^{n} c(i)x(i)$$

subject to:

$$\begin{aligned} \sum_{i=1}^{n} a(i,j)x(i) &= b(j), \quad j = 1, 2, \cdots, m \\ x(i) &\geq 0, \quad i = 1, 2, \cdots, n \end{aligned}$$

This primal LP has one variable $x(i)$ for each $i = 1$ to $n$. The data of the LP consists of numbers $c(i)$ for each $i = 1$ to $n$, the numbers $b(j)$ for each $j = 1$ to $m$ and the numbers $a(i, j)$ for each $i = 1$ to $n$ and for each $j = 1$ to $m$. The LP thus has $n$ variables and $m$ constraints. The dual of this LP is the following LP.

Variables: $y(j)$, $\forall j \in \{1, 2, \cdots, m'\}$

$$\text{Minimize} \quad \sum_{j=1}^{m'} b(j)y(j)$$

subject To:

$$\begin{aligned} \sum_{j=1}^{m'} a(i,j)y(j) &\geq c(i), \quad i = 1, 2, \cdots, n' \\ y(j) &\in (-\infty, +\infty), \quad j = 1, 2, \cdots, m' \end{aligned}$$

The dual LP has one variable $y(j)$ for each $j = 1$ to $m$. Each $y(j)$ variable is a **free** variable. That is, it is allowed to take any value in $\mathbb{R}$. The dual LP has $m$ variables and $n$ constraints.

The theorem of linear programming duality is as follows.

**Theorem B.1.** *(Luenberger, 1984) If either a primal LP or its dual LP has a finite optimal solution, then so does the other, and the corresponding values of the objective functions are equal.*

Applying this theorem to the primal-dual pair given above, there holds,

$$\sum_{i=1}^{n} c(i)x^*(i) = \sum_{j=1}^{m} b(j)y^*(j)$$

where $x^*$ denotes an optimal solution to the primal and $y^*$ denotes an optimal solution to the dual.

The theorem of complementary slackness is as follows.

**Theorem B.2.** *(Vanderbei, 2008) Suppose that $x$ is feasible for a primal linear program and $y$ is feasible for its dual. Let $(w_1,\cdots,w_m)$ denote the corresponding primal slack variables, and let $(z_1,\cdots,z_n)$ denote the corresponding dual slack variables. Then $x$ and $y$ are optimal for their respective problems if and only if*

$$\begin{aligned} x_j z_j = 0 \qquad &\text{for } j = 1, \cdots, n, \\ w_i y_i = 0 \qquad &\text{for } i = 1, \cdots, m. \end{aligned}$$

## Appendix C. Regret for DEC-POMDPs

The **value of an information set** $\varphi \in \mathcal{I}_i$ of an agent $i$ for a $i$-reduced joint policy $q$, denoted $\lambda_i^*(\varphi, q)$, is defined by:

$$\lambda_i^*(\varphi, q) \quad = \quad \max_{h \in \varphi} \sum_{j' \in \mathcal{E}_{-i}} \mathcal{R}(\alpha, \langle h, j' \rangle) q(j') \tag{160}$$

for any terminal information set and, if $\varphi$ is non-terminal, by:

$$\lambda_i^*(\varphi, q) \quad = \quad \max_{h \in \varphi} \sum_{o \in O_i} \lambda_i^*(h.o, q) \tag{161}$$

Then, the **regret of a history** $h$ for an agent $i$ and for a $i$-reduced joint policy $q$, denoted $\mu_i(h, q)$, it is defined by:

$$\mu_i(h, q) \quad = \quad \lambda_i^*(\varphi(h), q) - \sum_{j' \in H_{-i}^T} \mathcal{R}(\alpha, \langle h, j' \rangle) q(j') \tag{162}$$

if $h$ is terminal and, if $h$ is non-terminal, by:

$$\mu_i(h, q) \quad = \quad \lambda_i^*(\varphi(h), q) - \sum_{o \in O_i} \lambda_i^*(h.o, q) \tag{163}$$

The concept of regret of the agent $i$, which is independant of the policy of the agent $i$, is very useful when looking for optimal policy because its optimal value is known: it is 0. It is thus easier to manipulate than the optimal value of a policy.

## Appendix D. Program Changes Due to Optimizations

Pruning locally or globally extraneous histories reduces the size of the search space of the mathematical programs. Now, some constraints of the programs depend on the size of the search space, we must then alter some of these constraints.

Let denote by a "$\sim$" superscript the sets actually used in our program. For example, $\tilde{\mathcal{E}_i}$ will be the actual set of terminal histories of agent $i$, be it pruned of extraneous histories or not.

Programs **MILP** (Table 3) and **MILP-n agents** (Table 5) rely on the fact that the number of histories of a given length $t$ in the support of a pure policy of each agent is fixed and equal to $|O_i|^{t-1}$. As it may not be the case with pruned sets, the following changes have to be made:

- The constraint (42) of **MILP** or (121) **MILP-n agents**, that is

$$\sum_{j \in \mathcal{E}} z(j) \quad = \quad \prod_{i \in I} |O_i|^{T-1}$$

must be replaced by

$$\sum_{j \in \tilde{\mathcal{E}}} z(j) \quad \leq \quad \prod_{i \in I} |O_i|^{T-1}. \tag{164}$$

- The set of constraints (41) of **MILP** or (120) of **MILP-n agents**, that is

$$\sum_{j' \in \mathcal{E}_{-i}} z(\langle h, j' \rangle) \quad = \quad \prod_{k \in I \setminus \{i\}} |O_k|^{T-1} x_i(h), \quad \forall i \in I, \forall h \in \mathcal{E}_i$$

must be replaced by

$$\sum_{j' \in \tilde{\mathcal{E}}_{-i}} z(\langle h, j' \rangle) \quad \leq \quad \prod_{k \in I \setminus \{i\}} |O_k|^{T-1} x_i(h), \quad \forall i \in I, \forall h \in \tilde{\mathcal{E}}_i. \qquad (165)$$

- The set of constraints (119) of **MILP-n agents**, that is

$$y_i(\varphi(h)) - \frac{1}{|O_i|^{T-1}} \sum_{j \in \mathcal{E}} \mathcal{R}(\alpha, \langle h, j_{-i} \rangle) z(j) \quad = \quad w_i(h), \quad \forall h \in \mathcal{E}_i$$

must be replaced by

$$y_i(\varphi(h)) - \frac{1}{|\tilde{O}_i|^{T-1}} \sum_{j \in \tilde{\mathcal{E}}} \mathcal{R}(\alpha, \langle h, j_{-i} \rangle) z(j) \quad = \quad w_i(h), \quad \forall h \in \tilde{\mathcal{E}}_i. \qquad (166)$$

## Appendix E. Example using MA-Tiger

All these example are derived using the Decentralized Tiger Problem (MA-Tiger) described in Section 2.2. We have two agents, with 3 actions ($a_l$, $a_r$, $a_o$) and 2 observations ($o_l$, $o_r$). We will only consider problem with an horizon of 2.

There are 18 ($3^2 \times 2$) terminal histories for an agent: $a_o.o_l.a_o$, $a_o.o_l.a_l$, $a_o.o_l.a_r$, $a_o.o_r.a_o$, $a_o.o_r.a_l$, $a_o.o_r.a_r$, $a_l.o_l.a_o$, $a_l.o_l.a_l$, $a_l.o_l.a_r$, $a_l.o_r.a_o$, $a_l.o_r.a_l$, $a_l.o_r.a_r$, $a_r.o_l.a_o$, $a_r.o_l.a_l$, $a_r.o_l.a_r$, $a_r.o_r.a_o$, $a_r.o_r.a_l$, $a_r.o_r.a_r$.

And thus 324 ($18^2 = 3^{2\times2} \times 2^2$) joint histories for the agents: $\langle a_o.o_l.a_o, a_o.o_l.a_o \rangle$, $\langle a_o.o_l.a_o, a_o.o_l.a_l \rangle$, $\langle a_o.o_l.a_o, a_o.o_l.a_r \rangle$, $\cdots$, $\langle a_r.o_r.a_r, a_r.o_r.a_r \rangle$.

### E.1 Policy Constraints

The policy constraints with horizon 2 for one agent in the MA-Tiger problem would be:
Variables: $x$ for every history

$$\begin{aligned}
x(a_o) + x(a_l) + x(a_r) &= 0 \\
-x(a_o) + x(a_o.o_l.a_o) + x(a_o.o_l.a_l) + x(a_o.o_l.a_r) &= 0 \\
-x(a_o) + x(a_o.o_r.a_o) + x(a_o.o_r.a_l) + x(a_o.o_r.a_r) &= 0 \\
-x(a_l) + x(a_l.o_l.a_o) + x(a_l.o_l.a_l) + x(a_l.o_l.a_r) &= 0 \\
-x(a_l) + x(a_l.o_r.a_o) + x(a_l.o_r.a_l) + x(a_l.o_r.a_r) &= 0 \\
-x(a_r) + x(a_r.o_l.a_o) + x(a_r.o_l.a_l) + x(a_r.o_l.a_r) &= 0 \\
-x(a_r) + x(a_r.o_r.a_o) + x(a_r.o_r.a_l) + x(a_r.o_r.a_r) &= 0
\end{aligned}$$

$$\begin{array}{rrr} x(a_o) \geq 0 & x(a_l) \geq 0 & x(a_r) \geq 0 \\ x(a_o.o_l.a_o) \geq 0 & x(a_o.o_l.a_l) \geq 0 & x(a_o.o_l.a_r) \geq 0 \\ x(a_o.o_r.a_o) \geq 0 & x(a_o.o_r.a_l) \geq 0 & x(a_o.o_r.a_r) \geq 0 \\ x(a_l.o_l.a_o) \geq 0 & x(a_l.o_l.a_l) \geq 0 & x(a_l.o_l.a_r) \geq 0 \\ x(a_l.o_r.a_o) \geq 0 & x(a_l.o_r.a_l) \geq 0 & x(a_l.o_r.a_r) \geq 0 \\ x(a_r.o_l.a_o) \geq 0 & x(a_r.o_l.a_l) \geq 0 & x(a_r.o_l.a_r) \geq 0 \\ x(a_r.o_r.a_o) \geq 0 & x(a_r.o_r.a_l) \geq 0 & x(a_r.o_r.a_r) \geq 0 \end{array}$$

### E.2 Non-Linear Program for MA-Tiger

The Non-Linear Program for finding an optimal sequence-form policy for the MA-Tiger with horizon 2 would be:

Variables: $x_i$ for every history for each agent

$$\begin{array}{rl} \text{Maximize} & \mathcal{R}(\alpha, \langle a_o.o_l.a_o, a_o.o_l.a_o \rangle) x_1(a_o.o_l.a_o) x_2(a_o.o_l.a_o) \\ + & \mathcal{R}(\alpha, \langle a_o.o_l.a_o, a_o.o_l.a_l \rangle) x_1(a_o.o_l.a_o) x_2(a_o.o_l.a_l) \\ + & \mathcal{R}(\alpha, \langle a_o.o_l.a_o, a_o.o_l.a_r \rangle) x_1(a_o.o_l.a_o) x_2(a_o.o_l.a_r) \\ + & \cdots \end{array}$$

subject to:

$$\begin{array}{rcl} x_1(a_o) + x_1(a_l) + x_1(a_r) & = & 0 \\ -x_1(a_o) + x_1(a_o.o_l.a_o) + x_1(a_o.o_l.a_l) + x_1(a_o.o_l.a_r) & = & 0 \\ -x_1(a_o) + x_1(a_o.o_r.a_o) + x_1(a_o.o_r.a_l) + x_1(a_o.o_r.a_r) & = & 0 \\ -x_1(a_l) + x_1(a_l.o_l.a_o) + x_1(a_l.o_l.a_l) + x_1(a_l.o_l.a_r) & = & 0 \\ -x_1(a_l) + x_1(a_l.o_r.a_o) + x_1(a_l.o_r.a_l) + x_1(a_l.o_r.a_r) & = & 0 \\ -x_1(a_r) + x_1(a_r.o_l.a_o) + x_1(a_r.o_l.a_l) + x_1(a_r.o_l.a_r) & = & 0 \\ -x_1(a_r) + x_1(a_r.o_r.a_o) + x_1(a_r.o_r.a_l) + x_1(a_r.o_r.a_r) & = & 0 \end{array}$$

$$\begin{array}{rcl} x_2(a_o) + x_2(a_l) + x_2(a_r) & = & 0 \\ -x_2(a_o) + x_2(a_o.o_l.a_o) + x_2(a_o.o_l.a_l) + x_2(a_o.o_l.a_r) & = & 0 \\ -x_2(a_o) + x_2(a_o.o_r.a_o) + x_2(a_o.o_r.a_l) + x_2(a_o.o_r.a_r) & = & 0 \\ -x_2(a_l) + x_2(a_l.o_l.a_o) + x_2(a_l.o_l.a_l) + x_2(a_l.o_l.a_r) & = & 0 \\ -x_2(a_l) + x_2(a_l.o_r.a_o) + x_2(a_l.o_r.a_l) + x_2(a_l.o_r.a_r) & = & 0 \\ -x_2(a_r) + x_2(a_r.o_l.a_o) + x_2(a_r.o_l.a_l) + x_2(a_r.o_l.a_r) & = & 0 \\ -x_2(a_r) + x_2(a_r.o_r.a_o) + x_2(a_r.o_r.a_l) + x_2(a_r.o_r.a_r) & = & 0 \end{array}$$

$$\begin{array}{rrr}
x_1(a_o) \geq 0 & x_1(a_l) \geq 0 & x_1(a_r) \geq 0 \\
x_1(a_o.o_l.a_o) \geq 0 & x_1(a_o.o_l.a_l) \geq 0 & x_1(a_o.o_l.a_r) \geq 0 \\
x_1(a_o.o_r.a_o) \geq 0 & x_1(a_o.o_r.a_l) \geq 0 & x_1(a_o.o_r.a_r) \geq 0 \\
x_1(a_l.o_l.a_o) \geq 0 & x_1(a_l.o_l.a_l) \geq 0 & x_1(a_l.o_l.a_r) \geq 0 \\
x_1(a_l.o_r.a_o) \geq 0 & x_1(a_l.o_r.a_l) \geq 0 & x_1(a_l.o_r.a_r) \geq 0 \\
x_1(a_r.o_l.a_o) \geq 0 & x_1(a_r.o_l.a_l) \geq 0 & x_1(a_r.o_l.a_r) \geq 0 \\
x_1(a_r.o_r.a_o) \geq 0 & x_1(a_r.o_r.a_l) \geq 0 & x_1(a_r.o_r.a_r) \geq 0
\end{array}$$

$$\begin{array}{rrr}
x_2(a_o) \geq 0 & x_2(a_l) \geq 0 & x_2(a_r) \geq 0 \\
x_2(a_o.o_l.a_o) \geq 0 & x_2(a_o.o_l.a_l) \geq 0 & x_2(a_o.o_l.a_r) \geq 0 \\
x_2(a_o.o_r.a_o) \geq 0 & x_2(a_o.o_r.a_l) \geq 0 & x_2(a_o.o_r.a_r) \geq 0 \\
x_2(a_l.o_l.a_o) \geq 0 & x_2(a_l.o_l.a_l) \geq 0 & x_2(a_l.o_l.a_r) \geq 0 \\
x_2(a_l.o_r.a_o) \geq 0 & x_2(a_l.o_r.a_l) \geq 0 & x_2(a_l.o_r.a_r) \geq 0 \\
x_2(a_r.o_l.a_o) \geq 0 & x_2(a_r.o_l.a_l) \geq 0 & x_2(a_r.o_l.a_r) \geq 0 \\
x_2(a_r.o_r.a_o) \geq 0 & x_2(a_r.o_r.a_l) \geq 0 & x_2(a_r.o_r.a_r) \geq 0
\end{array}$$

### E.3 MILP for MA-Tiger

The **MILP** with horizon 2 for the agents in the MA-Tiger problem would be:
Variables:
$x_i(h)$ for every history of agent $i$
$z(j)$ for every terminal joint history

$$\begin{array}{lrl}
\text{Maximize} & & \mathcal{R}(\alpha, \langle a_o.o_l.a_o, a_o.o_l.a_o\rangle) z(\langle a_o.o_l.a_o, a_o.o_l.a_o\rangle) \\
& + & \mathcal{R}(\alpha, \langle a_o.o_l.a_o, a_o.o_l.a_l\rangle) z(\langle a_o.o_l.a_o, a_o.o_l.a_l\rangle) \\
& + & \mathcal{R}(\alpha, \langle a_o.o_l.a_o, a_o.o_l.a_r\rangle) z(\langle a_o.o_l.a_o, a_o.o_l.a_r\rangle) \\
& + & \cdots
\end{array}$$

subject to:

$$\begin{aligned}
x_1(a_o) + x_1(a_l) + x_1(a_r) &= 0 \\
-x_1(a_o) + x_1(a_o.o_l.a_o) + x_1(a_o.o_l.a_l) + x_1(a_o.o_l.a_r) &= 0 \\
-x_1(a_o) + x_1(a_o.o_r.a_o) + x_1(a_o.o_r.a_l) + x_1(a_o.o_r.a_r) &= 0 \\
&\dots \\
x_2(a_o) + x_2(a_l) + x_2(a_r) &= 0 \\
-x_2(a_o) + x_2(a_o.o_l.a_o) + x_2(a_o.o_l.a_l) + x_2(a_o.o_l.a_r) &= 0 \\
-x_2(a_o) + x_2(a_o.o_r.a_o) + x_2(a_o.o_r.a_l) + x_2(a_o.o_r.a_r) &= 0 \\
&\dots \\
z(\langle a_o.o_l.a_o, a_o.o_l.a_o\rangle) + z(\langle a_o.o_l.a_o, a_o.o_l.a_l\rangle) + z(\langle a_o.o_l.a_o, a_o.o_l.a_r\rangle) &= 2 \times x_1(a_o.o_l.a_o) \\
z(\langle a_o.o_l.a_o, a_o.o_l.a_o\rangle) + z(\langle a_o.o_l.a_l, a_o.o_l.a_o\rangle) + z(\langle a_o.o_l.a_r, a_o.o_l.a_o\rangle) &= 2 \times x_2(a_o.o_l.a_o) \\
z(\langle a_o.o_l.a_l, a_o.o_l.a_o\rangle) + z(\langle a_o.o_l.a_l, a_o.o_l.a_l\rangle) + z(\langle a_o.o_l.a_l, a_o.o_l.a_r\rangle) &= 2 \times x_1(a_o.o_l.a_l) \\
z(\langle a_o.o_l.a_o, a_o.o_l.a_l\rangle) + z(\langle a_o.o_l.a_l, a_o.o_l.a_l\rangle) + z(\langle a_o.o_l.a_r, a_o.o_l.a_l\rangle) &= 2 \times x_2(a_o.o_l.a_l) \\
&\dots
\end{aligned}$$

$$\begin{array}{rrr}
x_1(a_o) \geq 0 & x_1(a_l) \geq 0 & x_1(a_r) \geq 0 \\
x_1(a_o.o_l.a_o) \in \{0,1\} & x_1(a_o.o_l.a_l) \in \{0,1\} & x_1(a_o.o_l.a_r) \in \{0,1\} \\
x_1(a_o.o_r.a_o) \in \{0,1\} & x_1(a_o.o_r.a_l) \in \{0,1\} & x_1(a_o.o_r.a_r) \in \{0,1\} \\
\dots & & \\
x_2(a_o) \geq 0 & x_2(a_l) \geq 0 & x_2(a_r) \geq 0 \\
x_2(a_o.o_l.a_o) \in \{0,1\} & x_2(a_o.o_l.a_l) \in \{0,1\} & x_2(a_o.o_l.a_r) \in \{0,1\} \\
x_2(a_o.o_r.a_o) \in \{0,1\} & x_2(a_o.o_r.a_l) \in \{0,1\} & x_2(a_o.o_r.a_r) \in \{0,1\} \\
\dots & & \\
z(\langle a_o.o_l.a_o, a_o.o_l.a_o\rangle) \in \{0,1\} & z(\langle a_o.o_l.a_o, a_o.o_l.a_l\rangle) \in \{0,1\} & z(\langle a_o.o_l.a_o, a_o.o_l.a_r\rangle) \in \{0,1\} \\
z(\langle a_o.o_l.a_l, a_o.o_l.a_o\rangle) \in \{0,1\} & z(\langle a_o.o_l.a_l, a_o.o_l.a_l\rangle) \in \{0,1\} & z(\langle a_o.o_l.a_l, a_o.o_l.a_r\rangle) \in \{0,1\} \\
\dots & &
\end{array}$$

### E.4 MILP-2 Agents for MA-Tiger

The **MILP-2 agents** with horizon 2 for the agents in the MA-Tiger problem would be:
Variables:
$x_i(h)$, $w_i(h)$ and $b_i(h)$ for every history of agent $i$
$y_i(\varphi)$) for each agent and for every information set

$$\text{Maximize} \quad y_1(\varnothing)$$

subject to:

$$\begin{aligned}
x_1(a_o) + x_1(a_l) + x_1(a_r) &= 0 \\
-x_1(a_o) + x_1(a_o.o_l.a_o) + x_1(a_o.o_l.a_l) + x_1(a_o.o_l.a_r) &= 0 \\
-x_1(a_o) + x_1(a_o.o_r.a_o) + x_1(a_o.o_r.a_l) + x_1(a_o.o_r.a_r) &= 0 \\
\cdots \\
x_2(a_o) + x_2(a_l) + x_2(a_r) &= 0 \\
-x_2(a_o) + x_2(a_o.o_l.a_o) + x_2(a_o.o_l.a_l) + x_2(a_o.o_l.a_r) &= 0 \\
-x_2(a_o) + x_2(a_o.o_r.a_o) + x_2(a_o.o_r.a_l) + x_2(a_o.o_r.a_r) &= 0 \\
\cdots \\
y_1(\varnothing) - y_1(a_o.o_l) - y_1(a_o.o_r) &= w_1(a_o) \\
y_1(\varnothing) - y_1(a_l.o_l) - y_1(a_l.o_r) &= w_1(a_l) \\
y_1(\varnothing) - y_1(a_r.o_l) - y_1(a_r.o_r) &= w_1(a_r) \\
y_2(\varnothing) - y_2(a_o.o_l) - y_2(a_o.o_r) &= w_2(a_o) \\
y_2(\varnothing) - y_2(a_l.o_l) - y_2(a_l.o_r) &= w_2(a_l) \\
y_2(\varnothing) - y_2(a_r.o_l) - y_2(a_r.o_r) &= w_2(a_r)
\end{aligned}$$

$$\begin{aligned}
y_1(a_o.o_l) - \mathcal{R}(\alpha, \langle a_o.o_l.a_o, a_o.o_l.a_o\rangle)x_2(a_o.o_l.a_o) \\
-\mathcal{R}(\alpha, \langle a_o.o_l.a_o, a_o.o_l.a_l\rangle)x_2(a_o.o_l.a_l) \\
-\mathcal{R}(\alpha, \langle a_o.o_l.a_o, a_o.o_l.a_r\rangle)x_2(a_o.o_l.a_r) \\
-\mathcal{R}(\alpha, \langle a_o.o_l.a_o, a_l.o_l.a_o\rangle)x_2(a_l.o_l.a_o) \\
-\mathcal{R}(\alpha, \langle a_o.o_l.a_o, a_l.o_l.a_l\rangle)x_2(a_l.o_l.a_l) \\
-\mathcal{R}(\alpha, \langle a_o.o_l.a_o, a_l.o_l.a_r\rangle)x_2(a_l.o_l.a_r) \\
\cdots &= w_1(a_o.o_l.a_o)
\end{aligned}$$

$$\begin{aligned}
y_1(a_o.o_l) - \mathcal{R}(\alpha, \langle a_o.o_l.a_l, a_o.o_l.a_o\rangle)x_2(a_o.o_l.a_o) \\
-\mathcal{R}(\alpha, \langle a_o.o_l.a_l, a_o.o_l.a_l\rangle)x_2(a_o.o_l.a_l) \\
-\mathcal{R}(\alpha, \langle a_o.o_l.a_l, a_o.o_l.a_r\rangle)x_2(a_o.o_l.a_r) \\
-\mathcal{R}(\alpha, \langle a_o.o_l.a_l, a_l.o_l.a_o\rangle)x_2(a_l.o_l.a_o) \\
-\mathcal{R}(\alpha, \langle a_o.o_l.a_l, a_l.o_l.a_l\rangle)x_2(a_l.o_l.a_l) \\
-\mathcal{R}(\alpha, \langle a_o.o_l.a_l, a_l.o_l.a_r\rangle)x_2(a_l.o_l.a_r) \\
\cdots &= w_1(a_o.o_l.a_l)
\end{aligned}$$

$$\begin{aligned}
&& \cdots && \\
y_1(a_r.o_r) - \mathcal{R}(\alpha, \langle a_r.o_r.a_r, a_o.o_l.a_o\rangle)x_2(a_o.o_l.a_o) && && \\
-\mathcal{R}(\alpha, \langle a_r.o_r.a_r, a_o.o_l.a_l\rangle)x_2(a_o.o_l.a_l) && && \\
-\mathcal{R}(\alpha, \langle a_r.o_r.a_r, a_o.o_l.a_r\rangle)x_2(a_o.o_l.a_r) && && \\
-\mathcal{R}(\alpha, \langle a_r.o_r.a_r, a_l.o_l.a_o\rangle)x_2(a_l.o_l.a_o) && && \\
-\mathcal{R}(\alpha, \langle a_r.o_r.a_r, a_l.o_l.a_l\rangle)x_2(a_l.o_l.a_l) && && \\
-\mathcal{R}(\alpha, \langle a_r.o_r.a_r, a_l.o_l.a_r\rangle)x_2(a_l.o_l.a_r) && && \\
\cdots &&=&& w_1(a_r.o_r.a_r)
\end{aligned}$$

$$\begin{aligned}
y_2(a_o.o_l) - \mathcal{R}(\alpha, \langle a_o.o_l.a_o, a_o.o_l.a_o\rangle)x_1(a_o.o_l.a_o) && && \\
-\mathcal{R}(\alpha, \langle a_o.o_l.a_l, a_o.o_l.a_o\rangle)x_1(a_o.o_l.a_l) && && \\
-\mathcal{R}(\alpha, \langle a_o.o_l.a_r, a_o.o_l.a_o\rangle)x_1(a_o.o_l.a_r) && && \\
-\mathcal{R}(\alpha, \langle a_l.o_l.a_o, a_o.o_l.a_o\rangle)x_1(a_l.o_l.a_o) && && \\
-\mathcal{R}(\alpha, \langle a_l.o_l.a_l, a_o.o_l.a_o\rangle)x_1(a_l.o_l.a_l) && && \\
-\mathcal{R}(\alpha, \langle a_l.o_l.a_r, a_o.o_l.a_o\rangle)x_1(a_l.o_l.a_r) && && \\
\cdots &&=&& w_2(a_o.o_l.a_o)
\end{aligned}$$

$$\begin{aligned}
y_2(a_o.o_l) - \mathcal{R}(\alpha, \langle a_o.o_l.a_o, a_o.o_l.a_l\rangle)x_1(a_o.o_l.a_o) && && \\
-\mathcal{R}(\alpha, \langle a_o.o_l.a_l, a_o.o_l.a_l\rangle)x_1(a_o.o_l.a_l) && && \\
-\mathcal{R}(\alpha, \langle a_o.o_l.a_r, a_o.o_l.a_l\rangle)x_1(a_o.o_l.a_r) && && \\
-\mathcal{R}(\alpha, \langle a_l.o_l.a_o, a_o.o_l.a_l\rangle)x_1(a_l.o_l.a_o) && && \\
-\mathcal{R}(\alpha, \langle a_l.o_l.a_l, a_o.o_l.a_l\rangle)x_1(a_l.o_l.a_l) && && \\
-\mathcal{R}(\alpha, \langle a_l.o_l.a_r, a_o.o_l.a_l\rangle)x_1(a_l.o_l.a_r) && && \\
\cdots &&=&& w_2(a_o.o_l.a_l) \\
&& \cdots &&
\end{aligned}$$

$$\begin{aligned}
x_1(a_o) \le 1 - b_1(a_o) &\qquad x_1(a_l) \le 1 - b_1(a_l) \\
x_1(a_r) \le 1 - b_1(a_r) &\qquad x_1(a_o.o_l.a_o) \le 1 - b_1(a_o.o_l.a_o) \\
x_1(a_o.o_l.a_l) \le 1 - b_1(a_o.o_l.a_l) &\qquad x_1(a_o.o_l.a_r) \le 1 - b_1(a_o.o_l.a_r) \\
\cdots &
\end{aligned}$$

$$\begin{aligned}
w_1(a_o) \le \mathcal{U}_1(a_o)b_1(a_o) &\qquad w_1(a_l) \le \mathcal{U}_1(a_l)b_1(a_l) \\
w_1(a_r) \le \mathcal{U}_1(a_r)b_1(a_r) &\qquad w_1(a_o.o_l.a_o) \le \mathcal{U}_1(a_o.o_l.a_o)b_1(a_o.o_l.a_o) \\
w_1(a_o.o_l.a_l) \le \mathcal{U}_1(a_o.o_l.a_l)b_1(a_o.o_l.a_l) &\qquad w_1(a_o.o_l.a_r) \le \mathcal{U}_1(a_o.o_l.a_r)b_1(a_o.o_l.a_r) \\
\cdots &
\end{aligned}$$

$$
\begin{array}{rll}
x_1(a_o) \geq 0 & x_1(a_l) \geq 0 & x_1(a_r) \geq 0 \\
x_1(a_o.o_l.a_o) \geq 0 & x_1(a_o.o_l.a_l) \geq 0 & x_1(a_o.o_l.a_r) \geq 0 \\
\ldots & & \\
w_1(a_o) \geq 0 & w_1(a_l) \geq 0 & w_1(a_r) \geq 0 \\
w_1(a_o.o_l.a_o) \geq 0 & w_1(a_o.o_l.a_l) \geq 0 & w_1(a_o.o_l.a_r) \geq 0 \\
\ldots & & \\
b_1(a_o) \in \{0,1\} & b_1(a_l) \in \{0,1\} & b_1(a_r) \in \{0,1\} \\
b_1(a_o.o_l.a_o) \in \{0,1\} & b_1(a_o.o_l.a_l) \in \{0,1\} & b_1(a_o.o_l.a_r) \in \{0,1\} \\
\ldots & & \\
y_1(\emptyset) \in (-\infty,+\infty) & & \\
y_1(a_o.o_l) \in (-\infty,+\infty) & y_1(a_o.o_r) \in (-\infty,+\infty) & \\
\ldots & &
\end{array}
$$

... and the same for agent 2